%% file: anonymous-submission-latex-2023.tex
\documentclass[letterpaper]{article} 
\usepackage{aaai23}
\usepackage{times}  
\usepackage{helvet}  
\usepackage{courier}  
\usepackage[hyphens]{url}  
\usepackage{graphicx} 
\usepackage{natbib}  
\usepackage{caption} 
\usepackage{algorithm}
\usepackage{algorithmic}

\usepackage{xcolor}
\usepackage{subfigure}
\usepackage{tikz}
\usepackage{amsmath}
\usepackage{amsfonts}
\usepackage{paralist}
\usepackage{tabularx}
\usepackage{booktabs}
\usepackage{bbold}
\usepackage{placeins}
\usepackage{morefloats}
\usepackage{etoolbox}

\AtBeginEnvironment{tabular}{\small} 

\definecolor{red}{HTML}{E41A1C}
\definecolor{orange}{HTML}{FF7F00}
\definecolor{yellow}{HTML}{FFC020}
\definecolor{green}{HTML}{4DAF4A}
\definecolor{blue}{HTML}{377EB8}
\definecolor{purple}{HTML}{984EA3}

\usetikzlibrary{decorations.pathreplacing,calligraphy}

\newcommand{\yhat}{\hat{y}}

\newcommand{\fp}{\mathtt{fp}}
\newcommand{\fn}{\mathtt{fn}}

\newcommand{\false}{\mathtt{false}}

\newcommand{\bfx}{\mathbf{x}}

\newcommand{\bfy}{\mathbf{y}}
\newcommand{\bfyhat}{\hat{\mathbf{y}}}

\newcommand{\bfytilde}{\tilde{\mathbf{y}}}

\newcommand{\dataset}{\mathcal{D}}

\newcommand\blfootnote[1]{%
  \begingroup
  \renewcommand\thefootnote{}\footnote{#1}%
  \addtocounter{footnote}{-1}%
  \endgroup
}

\title{Query-Based Hard-Image Retrieval for Object Detection at Test Time}
\author{
    Edward Ayers\textsuperscript{*},
    Jonathan Sadeghi\textsuperscript{*},
    John Redford,
    Romain Mueller\textsuperscript{$\dagger$},
    Puneet K.~Dokania\textsuperscript{$\dagger$}
    \blfootnote{\textsuperscript{,$\dagger$} These authors contributed equally.}
}
\affiliations{
    Five AI Ltd., UK \\
    first.last@five.ai
}

\begin{document}

\maketitle

\begin{abstract}
There is a longstanding interest in capturing the error behaviour of object detectors by finding images where their performance is likely to be unsatisfactory.
In real-world applications such as autonomous driving, it is also crucial to characterise potential failures beyond simple requirements of detection performance.
For example, a missed detection of a pedestrian close to an ego vehicle will generally require closer inspection than a missed detection of a car in the distance. 
The problem of predicting such potential failures \textit{at test time} has largely been overlooked in the literature and conventional approaches based on detection uncertainty fall short in that they are agnostic to such fine-grained characterisation of errors.
In this work, we propose to reformulate the problem of finding ``hard'' images as a query-based hard image retrieval task, where queries are specific definitions of ``hardness'', and offer a simple and intuitive method that can solve this task for a large family of queries.
Our method is entirely post-hoc, does not require ground-truth annotations, is independent of the choice of a detector, and relies on an efficient Monte Carlo estimation that uses a simple stochastic model in place of the
ground-truth.
We show experimentally that it can be applied successfully to a wide variety of queries for which it can reliably identify hard images for a given detector without any labelled data. We provide results on ranking and classification tasks using the widely used RetinaNet, Faster-RCNN, Mask-RCNN, and Cascade Mask-RCNN object detectors. 
The code for this project is available at \url{https://github.com/fiveai/hardest}.
\end{abstract}

\begin{figure}[ht!]
\centering
\subfigure{\includegraphics[width=0.41\textwidth]{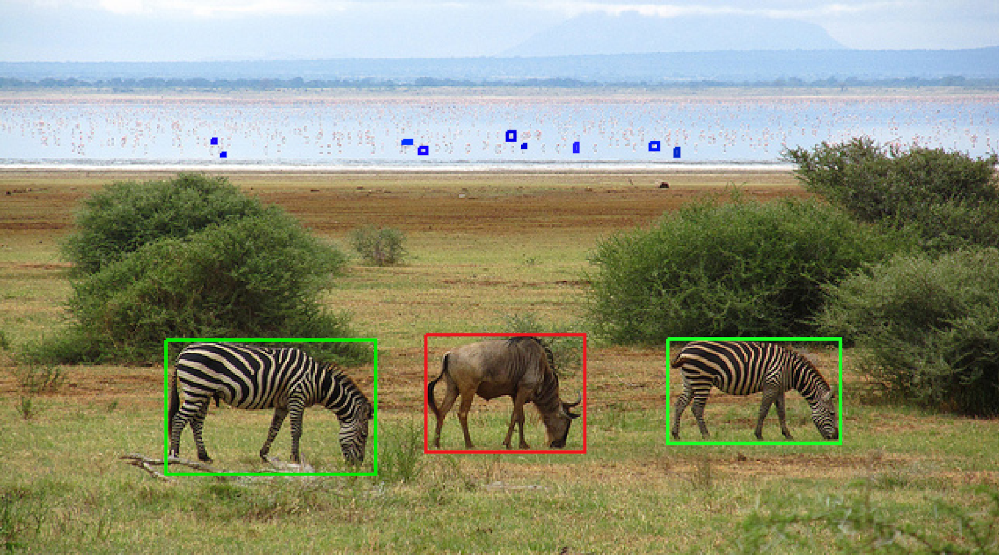}}
\subfigure{\includegraphics[width=0.41\textwidth]{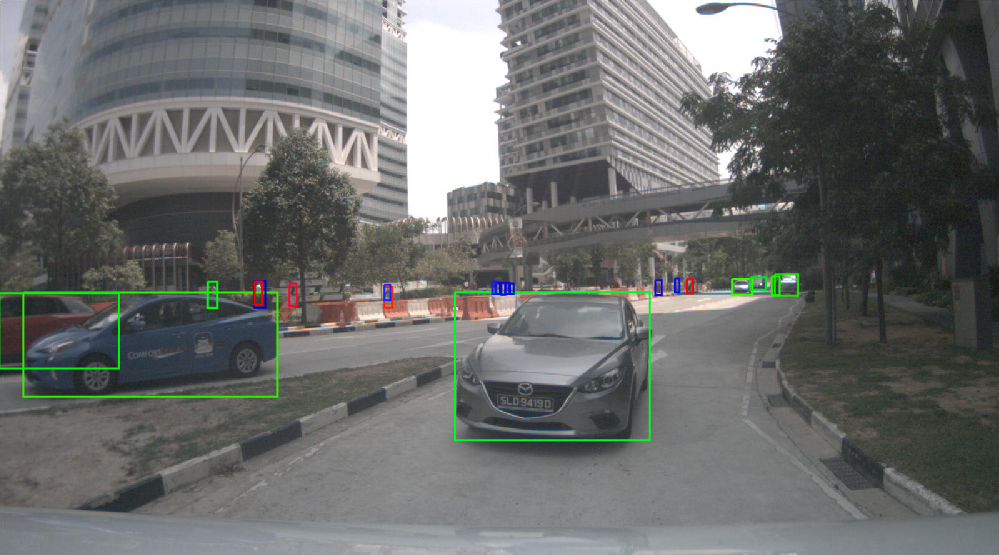}}
\caption{\label{fig:eg_flamingos}
Examples of images showing that ``hardness'' is task-specific for object detectors. Green, red, and blue boxes denote true positives, false positives, false negatives, respectively. (Top) Output of RetinaNet on an image from the COCO \emph{val} set.
Is this image hard for the detector because of the misclassified `wildebeest' in the foreground or because of the undetected `birds' in the background?
(Bottom) Output of Cascade-RCNN on a frame from the nuImages \emph{val} set.
Should this image be considered to be hard for this detector because of the undetected `pedestrians' even though they are further away compared to the correctly detected `cars'? Clearly, there is no single correct answer unless a specific requirement of performance is described.
}
\label{fig:introexamplefig}
\end{figure}

\section{Introduction}
\textit{How difficult (or hard) an image can be for a model at inference?} A large body of work has already paid significant attention in defining the notion of hard (or useful) images for standard classification and object detection tasks. The main objective of these works was to either perform active learning mainly to reduce the labelling and training cost \cite{Gal2017-vh,Choi2021-zg,Roy2018-lv, Aghdam2019-dx,Haussmann2020-vg,Yu2021-fe,Feng2019-yt}, or to improve the training mechanism~\cite{OHEM,retinanet,LibraRCNN,ovsampler}, perform metric learning~\cite{Suh2019-tf,Bucher2016-mp}, or even to improve the reliability of object detectors towards out-of-distribution samples~\cite{RegressionUncOD,VOS}. All these approaches attempt to improve the performance or robustness of object detectors by focusing on their training and generally assume that ground-truth labels are available. In contrast, our objective is to identify hard images for a given \textit{trained} detector, that is images for which the detector is likely to make a mistake, without having access to ground-truth labels.

First note that, in contrast to classification where there is a well-defined definition of error (a misclassification), there are multiple ways to define mistakes in object detection.
For example, an image can be considered hard if a detector produces many false-positives on it, but this by far not the only possibility. Another one is to say that an image is hard if the detector makes many mistakes on close objects rather than on distant ones. Therefore, what makes an image hard depends on the situation at hand and one should be able to quantify ``hardness'' accordingly by defining the associated notion of a mistake. We illustrate this point on Figure~\ref{fig:introexamplefig}, where we show two example images that exhibit a variety of possible definitions of errors for object detection.
Note that the widely used uncertainty metrics that are often used to quantify ``hardness'', such as entropy~\cite{ShannonEntropy} and Dempster-Shafer~\cite{sensoy2018evidential}, to name a few, are completely agnostic to such requirements. In this work, we revisit these notions and take a step towards defining and quantifying requirement-specific hardness metrics without having access to the ground-truth labels. 

Our main contributions are as follows. First, we provide a framework that allows to define a large family of requirement-specific hardness metrics that define a notion of error. We call them \textit{hardness queries}. The central idea is to provide fine-grained error behaviours of a given detector at test time (e.g., set of false-negative (FN) and false-positive (FP) bounding boxes) to allow users to compose a large family of complex queries. For example, if provided with the set of \textit{potential} false positive and false negative bounding boxes, a user may compose a query and define hardness as a metric that quantifies the mistake severity on these sets. Let us say that the user's query is to ``find images where the number of false-positive boxes strictly dominate the number of false-negative ones''. Then, the metric quantifying such query would simply be $\mathbb{1}_{|\text{FN}|>|\text{FP}|}$, where each image for which the indicator turns out to be one would be classified as the hard image under the the given query. In such a way, many queries, much more expressive than this one (discussed in the main paper), can be composed depending on the requirements of the application. 

Since, in the absence of ground-truth annotations, it is not possible to define the set of fine-grained errors mentioned above, we propose a simple approach to define a distribution of \textit{pseudo} ground-truth labels (a multivariate Bernoulli distribution) that is then used to define a distribution over these error sets. This distribution is then employed via an efficient Monte Carlo sampling procedure to estimate the expected query-based hardness per image. 
Through extensive experiments using a suite of widely used object detectors (RetinaNet, Faster-RCNN, Mask-RCNN, and Cascade Mask-RCNN), we show that our method provides state-of-the-art performance (with significant margins in many cases) on ranking and classification tasks that tackle the problem of finding query-based hard images at test time.


\section{Defining Query-Based Hardness}
\label{sec:problem_formulation}
Let us denote a general object detector as a function $D_\eta$ that maps an input image $\bfx$ to a list of \emph{detection instances}\footnote{Here $\bfyhat$ denotes the \emph{final} output of the detector (e.g. after NMS or any other post-processing) and $\eta$ is typically chosen according to the task-specific requirements for tradeoffs of precision and recall.} $\bfyhat = D_\eta(\bfx)$, where each instance $\yhat_i = (b_i, c_i, s_i)$ consists of a regressed bounding box $b_i$ enclosing the $i$-th detected object, a class label $c_i \in \{1, \cdots, K\}$, and a detection score $s_i \in [0, 1]$.
Here $\eta$ is the threshold such that only instances with $s_i \geq \eta$ are produced by the detector $D_\eta$.
We denote the ground-truth instances by $\bfy = \{y_i: i = 1, \ldots \}$, where each instance is of the form $y_i = (b_i, c_i)$.
We do not make any further assumptions regarding the architecture or the training procedure of the detector.

\subsection{Image Hardness the Conventional Way}
\label{sec:uncertainy}
A widely accepted approach to capture how difficult or hard a certain input is for a given classification model is to compute the entropy of its predictive distribution~\cite{Gal2017-vh}. The concept can be extended to object detectors as well~\cite{Roy2018-lv}. Assuming that we have access to the full probability vector $\mathbf p_i \in [0,1]^K$ over the $K$ classes for each bounding box, we can define the total \emph{entropy} per image as
\begin{align} \label{eq:entropy}
    \mathrm{H}_{\bfx}(\bfyhat) = -\sum_{i=1}^{|\bfyhat|} \sum_{k=1}^K p_i(k) \log p_i(k),
\end{align}
where we sum over all the detected instances $i$, and $p_i(k)$ is the probability with which the $i$-th bounding box belongs to the $k$-th category.
Eq.~\eqref{eq:entropy} captures the total uncertainty associated with an image as the sum of the entropy for the predicted bounding boxes.
This can be easily adapted to one-vs-all binary setups in which case $p_i(k)$ are independent binary probabilities for each class.

Recently, the Dempster-Shafer framework~\cite{Dempster2008, sensoy2018evidential} received significant attention in image classification tasks and has been shown to outperform entropy in estimating uncertainty for certain out-of-distribution detection and domain shift problems~\cite{Liu2020-sngp,PintoRegMixup2022}.
Contrary to the entropy, this metric is defined in terms of the \emph{evidence} per class (also known as belief) instead of their probabilities, which is defined as $e_i(k) = \exp{l_i(k)}$, where $\mathbf l_i$ is the logit corresponding to $\mathbf p_i$.
For object detectors, the per-image Dempster-Shafer theory based evidential uncertainty can then be computed as:
\begin{align} \label{eq:DS}
    \mathrm{DS}_{\bfx}(\bfyhat) = \sum_{i=1}^{|\bfyhat|} \frac{K}{K + \sum_k e_i(k)},
\end{align}
where, similarly to $\mathrm{H}_{\bfx}(\bfyhat)$,  we sum over all the detected instances $i$ in the image.

\paragraph{Limitations}
Though both $\mathrm{H}_{\bfx}(\bfyhat)$ and $\mathrm{DS}_{\bfx}(\bfyhat)$ can be used at inference to quantify how likely it is that the given image is hard for the detector, using such measures of uncertainty to quantify ``hardness'' is crude as it does not allow one to specify arbitrary requirements of performance.
In real-world applications, it is crucial to have a more fined-grained characterisation of detection errors since different types of errors can have vastly different consequences for the downstream task (see Figure~\ref{fig:eg_flamingos} for an illustration).
Both $\mathrm{H}_{\bfx}(\bfyhat)$ and $\mathrm{DS}_{\bfx}(\bfyhat)$ do not allow specifying such requirements and hence suffer from the fact that they are \textit{agnostic} to the specific performance measure under consideration.

\subsection{Query-Based Hardness} \label{sec:task definition}

We now introduce a simple framework that allows to define a large family of hardness metrics, which we denote as \emph{hardness queries} in what follows.
Let us assume for now that we have access to the ground-truth bounding boxes $\bfy$ for each image $\bfx$.
This requirement will obviously not be fulfilled at test time but we present in the next section a method that can approximately resolve this limitation.
Given detection instances $\bfyhat = D_{\eta}(\bfx)$ and ground-truth bounding boxes $\bfy$, we can use the Hungarian algorithm on a thresholded intersection-over-union cost matrix~\cite{forsyth2012computer} to define the error sets $\mathtt{e}(\bfyhat, \bfy) \in \{\fp(\bfyhat, \bfy),\allowbreak \fn(\bfyhat, \bfy),\allowbreak \false(\bfyhat, \bfy)\}$, where $\fp$ and $\fn$ denote the set of false-positive and false-negative bounding boxes, respectively, and $\false = \fp \cup \fn$. 
The set of false-negative boxes consists of those ground-truth instances that have not been associated to any detection.

A large family of domain-specific hardness queries capturing the error characteristics of a detector can be introduced using the above fine-grained information about the error categories of the detected boxes.
For example, one could consider the relative position of the false-negatives and false-positives to be important when defining hardness, or might want to pay more attention to a particular class (for example, pedestrians) and define hardness accordingly.
We introduce below a few query-specific hardness definitions and use them throughout our experiments. 
However, depending on the task, more expressive queries can easily be constructed.

\paragraph{Examples of hardness queries}
In what follows, $\mathtt{e}$ denotes one of the error sets $\fp$, $\fn$, or $\false$ of the detected instances on an image for a specific detector. For each $\mathtt e$, we define the following hardness queries:
\begin{enumerate}
    \item[{\em Total number of errors}] An image $\bfx$ is considered harder than $\bfx'$ if the detector makes a larger number of errors of a specific type $\mathtt e$ on that image, that is if $|\mathtt{e}(\bfyhat, \bfy)| > |\mathtt{e}(\bfyhat', \bfy')|$.
    The corresponding query is defined by $\texttt{Total}_{\bfx} (\mathtt{e}(\bfyhat, \bfy)) = |\mathtt{e}(\bfyhat, \bfy)|$.
    \item[{\em Pixel-adjusted errors}] Here we consider errors on large objects to be more severe than errors on the smaller ones, and introduce a query that captures this requirement. To do so, we simply weight each error by the size of the corresponding bounding box as a fraction of the image area, and define
    \begin{align}
    \label{eq:pixeladjusted}
        \texttt{PixelAdj}_{\bfx}(\mathtt e(\bfyhat, \bfy)) = \sum_{b \in \mathtt e(\bfyhat, \bfy)} \frac{\texttt{area}(b)}{\texttt{area}(\bfx)},
    \end{align}
    where $\texttt{area}(.)$ denotes the area of the image or the bounding box.
    \item[{\em Occlusion-aware errors}] Here we consider images with many occluded objects to be harder than the less cluttered ones. Since we do not have 3D information, we make the reasonable assumption that overlaps with true positive boxes is a good proxy for measuring occlusions.
    Similarly to the pixel-adjusted case, we quantify  occlusion-aware hardness as:
    \begin{multline}
        \texttt{OccAware}_\bfx (\mathtt e(\bfyhat, \bfy))= \sum_{\substack{b \in \mathtt e(\bfyhat, \bfy)\\ b' \in \mathtt{tp}(\bfx)}} \frac{\texttt{inter}(b, b')}{\texttt{area}(b)}, 
    \end{multline}
    where, $\texttt{inter}(b, b')$ is the area of the intersection between $b$ and $b'$.
\end{enumerate}
In the appendix, we show on Figure~10 that the Spearman's rank correlation coefficient between all of the proposed hardness queries varies widely, indicating that they are indeed able to capture different notions of hardness.

\section{Quantifying Query-Based Hardness}

In the previous section, we assumed knowledge of the ground-truth bounding boxes in order to define application-specific notions of hardness.
In what follows, we show that this limitation can be circumvented by defining a \emph{distribution} over possible ground-truths given detected instances (we call them pseudo ground-truths) and that this distribution can be used to estimate the hardness of an image. 

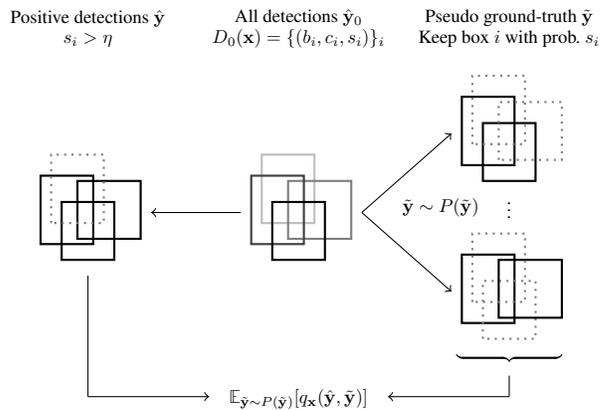
\begin{figure}[t!]
    \centering
    \begin{tikzpicture}[scale=0.7, every node/.style={scale=0.7}]
\begin{scope}[shift={(4cm,0cm)}]
\begin{scope}[shift={(4cm,0cm)}]
\draw[lightgray, thick] (0.2,0.4) rectangle (1+0.2,1.3+0.4);
\draw[gray, thick] (0.7,0.1) rectangle (1.2+0.7,1.1+0.1);
\draw[darkgray, thick] (0,0) rectangle (1,1.3);
\draw[black, thick] (0.4,-0.3) rectangle (1+0.4,1.1-0.3);
\node[align=center] at (0.9,4.1) {All detections $\bfyhat_0$\\$D_0(\bfx) = \{(b_i, c_i, s_i)\}_i$};
\end{scope}
\begin{scope}[shift={(0cm,0cm)}]
\draw[black, thick] (0,0) rectangle (1,1.3);
\draw[black, thick] (0.4,-0.3) rectangle (1+0.4,1.1-0.3);
\draw[black, thick] (0.7,0.1) rectangle (1.2+0.7,1.1+0.1);
\draw[gray, thick, dotted] (0.2,0.4) rectangle (1+0.2,1.3+0.4);
\node[align=center] at (0.9,4.1) {Positive detections $\bfyhat$\\$s_i > \eta$};
\end{scope}
\begin{scope}[shift={(8cm,0cm)}]
\begin{scope}[shift={(0cm,1.5cm)}]
\draw[black, thick] (0,0) rectangle (1,1.3);
\draw[black, thick] (0.4,-0.3) rectangle (1+0.4,1.1-0.3);
\draw[gray, thick, dotted] (0.7,0.1) rectangle (1.2+0.7,1.1+0.1);
\draw[gray, thick, dotted] (0.2,0.4) rectangle (1+0.2,1.3+0.4);
\end{scope}
\node[align=center] at (-0.4,0.7) {$\bfytilde \sim P(\bfytilde)$};
\node[align=center] at (0.9,0.75) {\vdots};
\begin{scope}[shift={(0cm,-1.5cm)}]
\draw[black, thick] (0,0) rectangle (1,1.3);
\draw[gray, thick, dotted] (0.4,-0.3) rectangle (1+0.4,1.1-0.3);
\draw[black, thick] (0.7,0.1) rectangle (1.2+0.7,1.1+0.1);
\draw[gray, thick, dotted] (0.2,0.4) rectangle (1+0.2,1.3+0.4);
\end{scope}
\node[align=center] at (0.9,4.1) {Pseudo ground-truth $\bfytilde$\\Keep box $i$ with prob. $s_i$};
\end{scope}
\draw[->] (3.8, 0.6) to (2.1, 0.6);
\draw[->] (6.1, 0.6) to (7.8, 2.1);
\draw[->] (6.1, 0.6) to (7.8,-0.9);
\begin{scope}[shift={(0cm,-1.9cm)}]
\draw[-] (0.9, 1.3) to (0.9,-1);
\draw[->] (0.9,-1) to (3.2, -1);
\draw[-] (8.9, -0.6) to (8.9,-1);
\draw[->] (8.9, -1) to (6.6,-1);
\draw[decorate, decoration = {calligraphic brace}, thick] (9.9, -0.2) --  (7.9, -0.2);
\node[align=center] at (4.9,-1.05) {$\mathbb E_{\bfytilde\sim P(\bfytilde)}[q_{\bfx} (\bfyhat, \bfytilde)]$};
\end{scope}
\end{scope}
\end{tikzpicture}

    \caption{\label{fig:score_sampling}
    Illustration of our method: starting from all detections ($\eta=0$) we generate pseudo-ground truth annotations $\bfytilde$ by selecting each box with probability given by their detection score $s_i$ and use them to evaluate positive detections $\bfyhat$ ($s_i>\eta$) in place of the ground truth using the hardness query $q_\bfx$, see Algorithm~\ref{algo:score_sampling} for details.}
\end{figure}

Our main insight is to treat the detection scores of the detected instances as posterior predictive probabilities and use samples from their implied distribution to estimate a given hardness query.
More precisely, for a given image $\bfx$, we first obtain the set of \emph{all detected instances} $\bfyhat_0 = D_0(\bfx)$.\footnote{In practice, we do not set $\eta$ to be exactly zero but rather choose a very small value. } This includes detections with $s_i < \eta$ which would appear as false negatives in the filtered detector output. For each detected instance $\yhat_i = (b_i, c_i, s_i ) \in \bfyhat_0$, we then introduce a Bernoulli distributed random variable $X_i \sim \text{Bernoulli}(s_i)$ parameterised by the corresponding detection score $s_i$. Therefore, if $|\bfyhat_0| = m$, we construct a multivariate Bernoulli distribution with $m$ independent variables.
We finally generate pseudo ground-truth instances $\bfytilde$ by selecting each detection $\yhat_i \in \bfyhat_0$ independently according to its Bernoulli distribution, i.e.~if $X_i = 1$.
This generates a distribution $P(\bfytilde)$ over pseudo ground-truth instances.

Denoting by $q_{\bfx}(\bfyhat, \bfytilde)$ a generic hardness query (such as Eq.~\ref{eq:pixeladjusted}), we estimate the hardness of an image by computing the expectation of the corresponding query over the pseudo ground-truth using the following Monte Carlo estimator:
\begin{align}
\label{eq:finalhardness}
    \mathrm{SS}_\bfx (\bfyhat; q) &= \mathbb E_{\bfytilde\sim P(\bfytilde)}[q_{\bfx} (\bfyhat, \bfytilde)] = \int q_{\bfx}(\bfyhat, \bfytilde) \mathrm d P(\bfytilde) \nonumber \\
    & \approx \frac{1}{N} \sum_{n=1}^N q_{\bfx}(\bfyhat, \bfytilde^{(n)}), \quad \bfytilde^{(n)} \sim P(\bfytilde),
\end{align}
where $N$ is the number of Monte Carlo samples.
We term this methodology \emph{score sampling} and summarize it in Algorithm~\ref{algo:score_sampling}.
Sampling $\bfytilde \sim P(\bfytilde)$ is extremely efficient as it typically requires a low number (typically $\approx 10$ in our experiments) of parallelizable Bernoulli trials.
An illustration of our approach is given in Figure~\ref{fig:score_sampling}.

A key assumption in our approach is that a detector will generally assign a low probability to ambiguous (or incorrectly classified) objects.
Therefore, even if the assigned probability of such an object is above the detection threshold, using this probability as the parameter of the Bernoulli distribution will likely lead to multiple unsuccessful trials and that bounding box will be included in the pseudo ground-truth only some of the time.
Conversely, bounding boxes with a high detection score will be sampled more frequently and be included in the pseudo ground-truth most of the time.
This assumption falls short when a detector is wrong with a very high confidence.
Since modern detectors are highly accurate and also show good calibration properties (refer to Figure~5 in the appendix for confidence histograms), we do not expect this assumption to fail frequently. An example would be RetinaNet~\cite{retinanet} which uses focal loss that has been shown to provide good calibration properties for image classification tasks~\cite{mukhoti2020calibrating}.
However, as a future work, it would be interesting to explore the impact of varying the calibration of object detectors on our approach.


 \begin{algorithm}[tb]
 \caption{Score Sampling (SS)} \label{algo:score_sampling}
\textbf{Input}: Image $\bfx$, detector $D_{\eta}$, hardness query $q_\bfx(\bfyhat, \bfy)$. \\
\textbf{Parameter}: Number of samples $N$.\\
\textbf{Output}: Estimated hardness $\hat q$ using score sampling.
\begin{algorithmic}[1] 
\STATE $\bfyhat_0 \leftarrow  D_0(\bfx)$
\FOR{$n \in \{1, ..., N\}$}
\FOR{$( b_i, c_i, s_i ) \in \bfyhat_0$}
\STATE $X_i \sim \text{Bernoulli}(s_i)$
\ENDFOR
\STATE $\bfytilde^{(n)} \gets \{( b_i, c_i) \text{ for } ( b_i, c_i, s_i) \in \bfyhat_0 |  X_i = 1\}$
\ENDFOR
\STATE $\bfyhat \leftarrow  \{( b_i, c_i, s_i) \in \bfyhat_0 | s_i > \eta \}$ 
\STATE $\hat q \gets \frac{1}{N} \sum_{n=1}^N q_\bfx(\bfyhat, \bfytilde^{(n)})$ 
\STATE \textbf{return} $\bar q $
\end{algorithmic}
\end{algorithm}

\section{Experiments}
\label{sec:evaluation}

We now illustrate the effectiveness of our approach in a variety of settings.
The following assumes a hardness query $q$ and a given detector $D_\eta$ at a given score threshold $\eta$ but we generally omit references to it for the ease of notation.
Given an annotated dataset $\dataset = \{(\bfx_i, \bfy_i): i=1, \ldots, D\}$, we denote by $q_i = q_{\bfx_i} (\bfyhat_i, \bfy_i)$ the \textit{ground-truth} hardness score of the $i$-th image obtained using the ground-truth bounding boxes $\bfy_i$, and by $\hat q_i$ the hardness score estimated by one of the methods under consideration.
In the case of \textit{score sampling (our)}, it is the approximation to the expectation $\mathbb E_{\bfytilde\sim P(\bfytilde)}[q_{\bfx_i} (\bfyhat_i, \bfytilde_i)]$ as defined in Eq.~\ref{eq:finalhardness}, while for the baselines, it is either computed using the entropy or the DS.

\paragraph{Datasets and detectors}
To illustrate the generality of our method, we evaluate off-the-shelf models on public-domain datasets for which weights are readily available.
We perform our evaluations on the the COCO dataset \cite{lin2014microsoft} and nuImages~\cite{nuscenes2019} and consider the following detectors:
\begin{compactitem}
    \item \emph{coco-retina} and \emph{coco-rcnn} are RetinaNet~\cite{retinanet} and Faster-RCNN~\cite{ren2015faster} with a ResNet-50-FPN backbone trained on the COCO dataset, from \texttt{torchvision}~\cite{paszke2019pytorch}.
    \item \emph{mmdet-maskrcnn} and \emph{mmdet-cascade} are Mask-RCNN~\cite{he2017mask} and Cascade Mask-RCNN~\cite{cai2019cascade} with a ResNet-50-FPN backbone trained on nuImages, from \texttt{mmdet}~\cite{mmdet3d}.
    We dismiss all instance masks in our experiments and only consider bounding boxes.
    We remap the nuScenes and MMDetection label schemas to a simplified two-class schema (pedestrian and vehicle) which we describe in the appendix.
\end{compactitem}
We set the pre-NMS score threshold to $\eta = 0.05$ for all detectors.
We perform our evaluations on the \emph{val} sets of the COCO and nuImages datasets.

\paragraph{Query-based Hardness}
Throughout the experiments, we use the queries introduced above but our evaluation protocol can be extended to a large family of single-image measure of performance.
More precisely, we will consider all combinations of the counting, pixel-adjusted, and occlusion-aware queries applied to the error categories $\texttt{fp}$, $\texttt{fn}$, and $\texttt{false} = \texttt{fp} \cup \texttt{fn}$, giving us 9 hardness queries in total.

\paragraph{Baselines} We consider entropy (Eq.~\ref{eq:entropy}) and Dempster-Shafer (Eq.~\ref{eq:DS}) uncertainty estimates as our baselines.
Both methods measure the uncertainty associated with an image using a categorical distribution over all possible classes for each detected box.
For two-stage architectures (Faster-RCNN, Mask-RCNN, and Cascade-RCNN), we modify the standard implementation to get access to the set of $K$ logits for each box which are then normalised to a categorical distribution with probabilities $\mathbf p$ using a softmax.
In this case, we use use Eq.~\eqref{eq:entropy} and \eqref{eq:DS} directly with $K$ being the total number of classes (including the background class).
The situation is different for RetinaNet, in which each box gets instead $K$ independent binary scores $\mathbf p \in [0, 1]^{K}$, representing the one-vs-all probability for that class. It is not clear in this case how to define a predictive categorical distribution over all classes and we instead treat this situation as a binary problem with $K=2$ and compute the uncertainty estimates for the detection  score directly. 

\paragraph{Implementation details}
We use \texttt{pycocotools}~\cite{lin2014microsoft} for processing bounding boxes, including for computing the association.
We use an IOU threshold of 0.5 for the association of bounding boxes in \texttt{pycocotools}.
We compute score sampling (Algorithm~\ref{algo:score_sampling}) using 10 Monte Carlo samples in all experiments.
This provides a favourable balance between accuracy and computational expense, and we provide a sensitivity analysis of the number of samples in the appendix.

\subsection{Qualitative Results}
\label{sec:qualitative}

\begin{figure*}
\begin{tabularx}{\textwidth}{ c c c c }
\toprule
   & \multicolumn{3}{c}{\textbf{Query 1}: Total False Positives ($\texttt{Total}(\texttt{fp})$)} \\
 \cmidrule{2-4} 
 \raisebox{0.6cm}{\textbf{SS (our)}} & \includegraphics{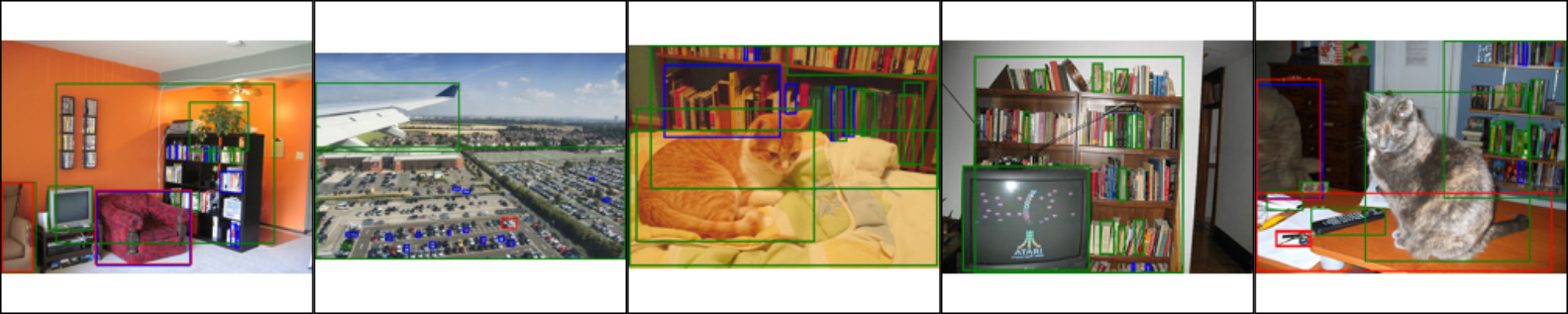} & \raisebox{0.6cm}{$\cdots$} &\includegraphics{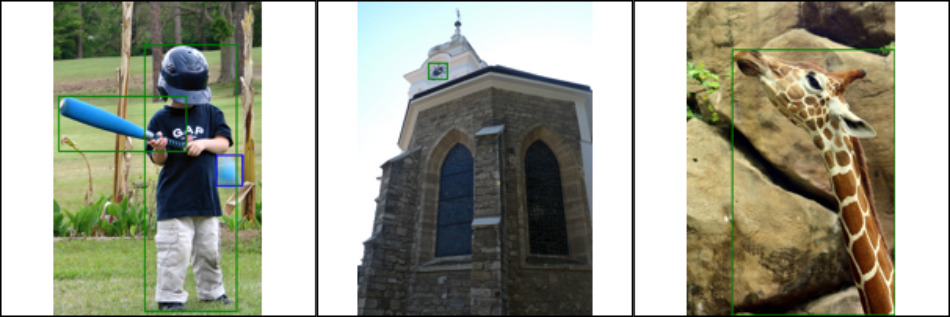} \\
 \raisebox{0.6cm}{\textbf{GT Ranking}} & \includegraphics{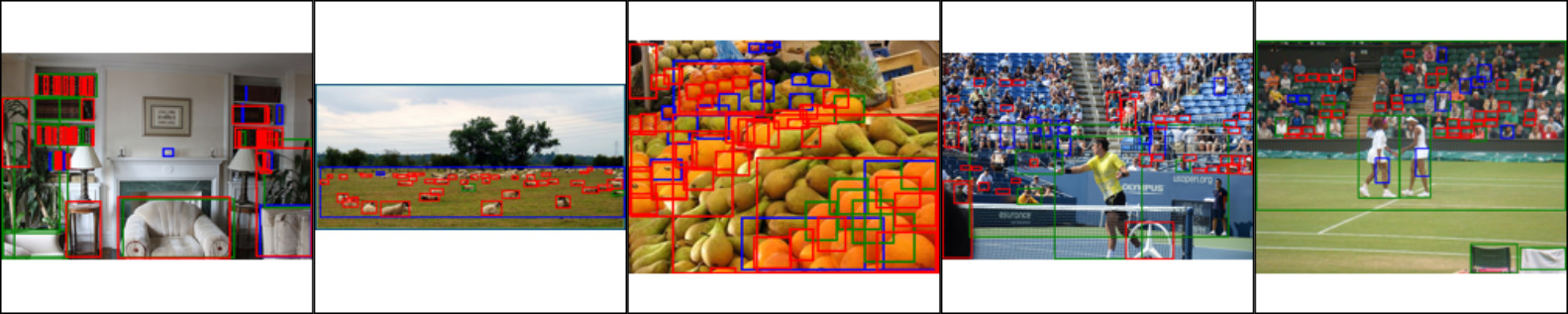} & \raisebox{0.6cm}{$\cdots$} &\includegraphics{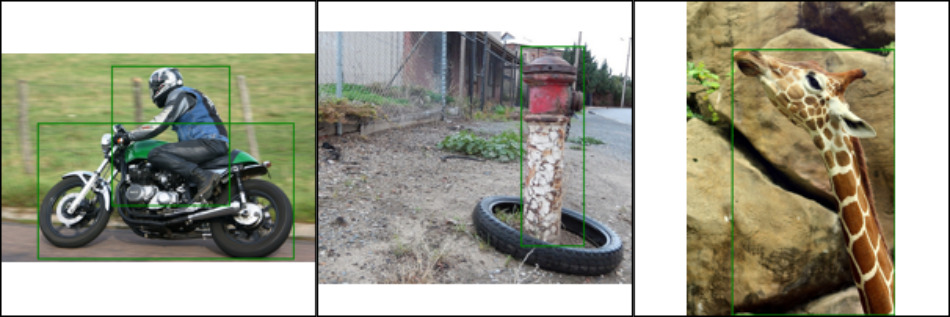} \\
 \cmidrule{2-4} 
   & \multicolumn{3}{c}{\textbf{Query 2}: Occlusion-aware False Positives ($\texttt{OccAware}(\texttt{fp})$)} \\
 \cmidrule{2-4} 
 \raisebox{0.6cm}{\textbf{SS (our)}} &  \includegraphics{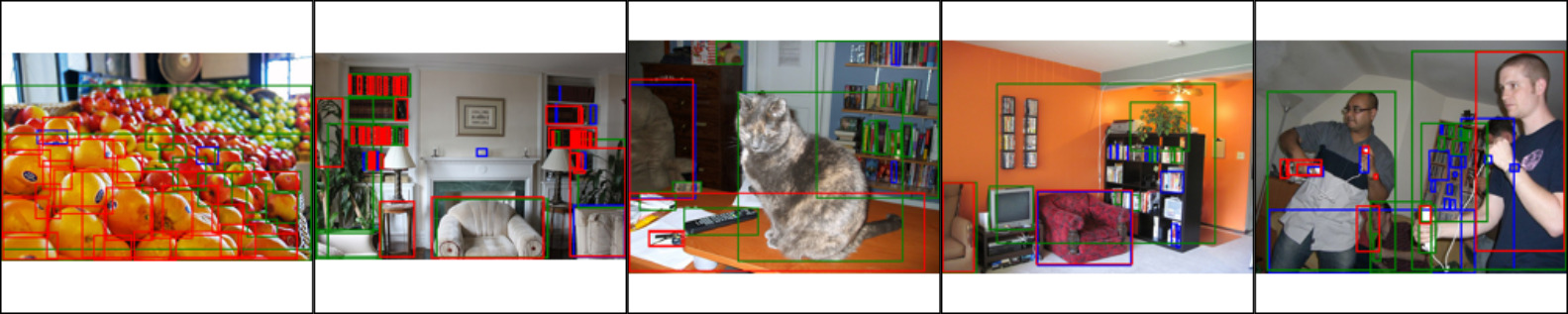} & \raisebox{0.6cm}{$\cdots$} &\includegraphics{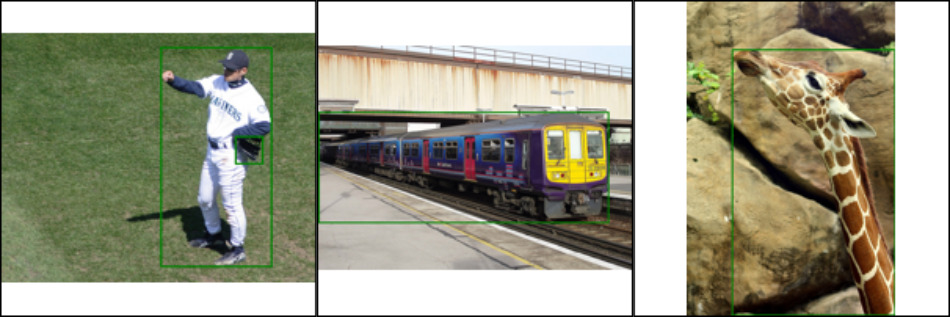} \\
 \raisebox{0.6cm}{\textbf{GT Ranking}} & \includegraphics{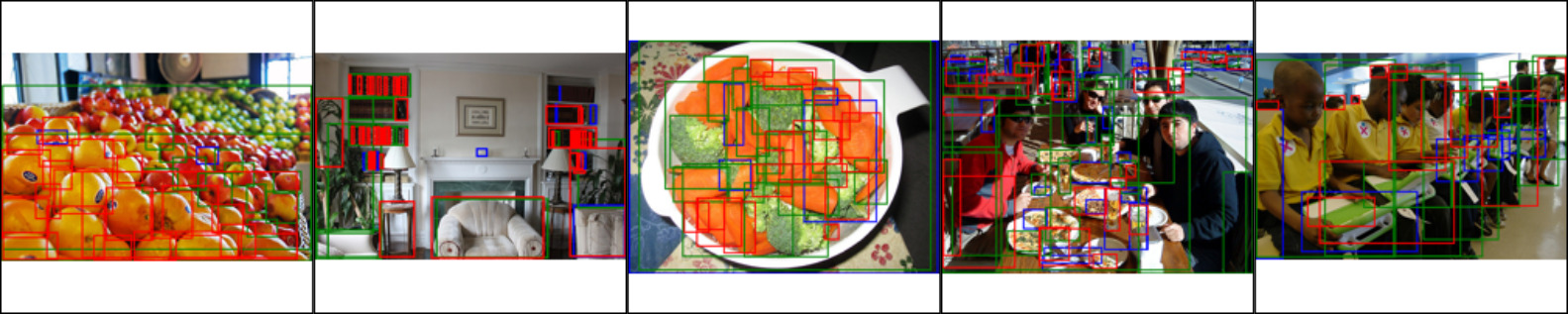} & \raisebox{0.6cm}{$\cdots$} &\includegraphics{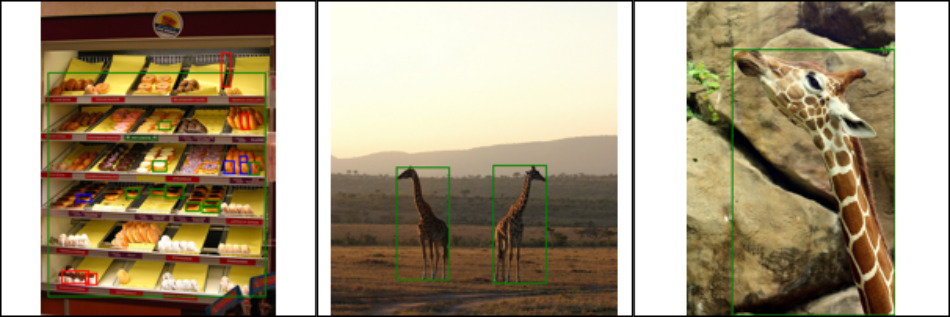} \\
\cmidrule{2-4} 
   & \multicolumn{3}{c}{\textbf{Query 3}: Pixel-adjusted False Positives ($\texttt{PixelAdj}(\texttt{fp})$)} \\
 \cmidrule{2-4} 
 \raisebox{0.6cm}{\textbf{SS (our)}} & \includegraphics{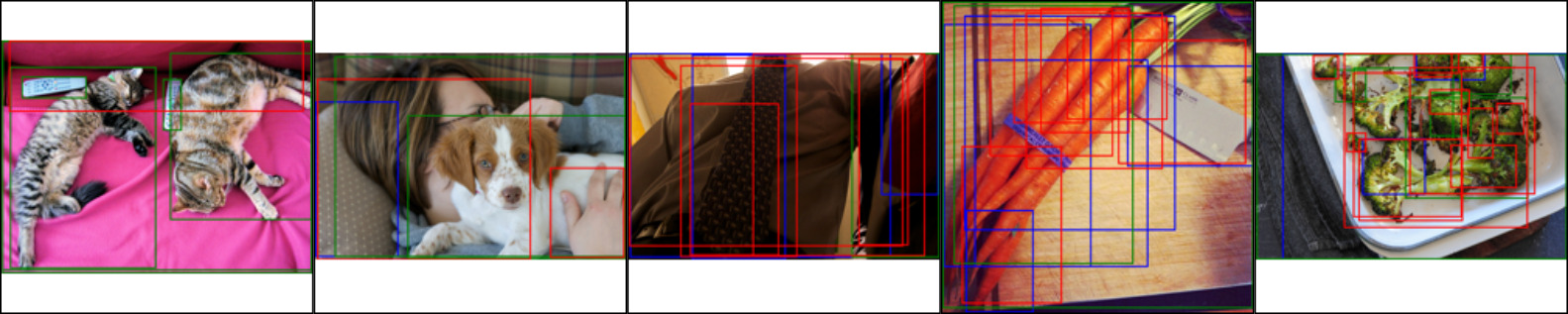} & \raisebox{0.6cm}{$\cdots$} &\includegraphics{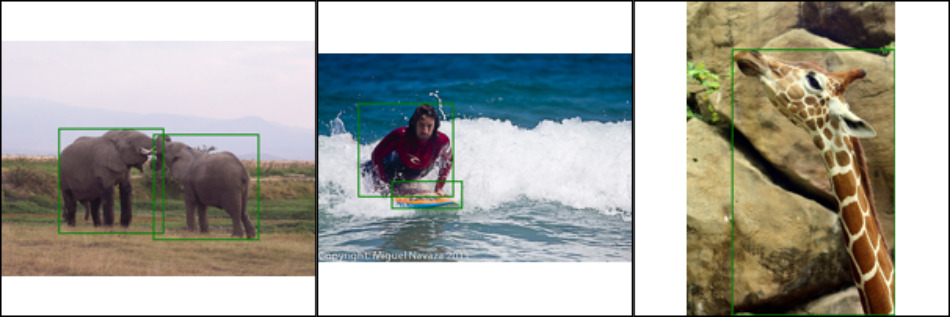} \\
 \raisebox{0.6cm}{\textbf{GT Ranking}} & \includegraphics{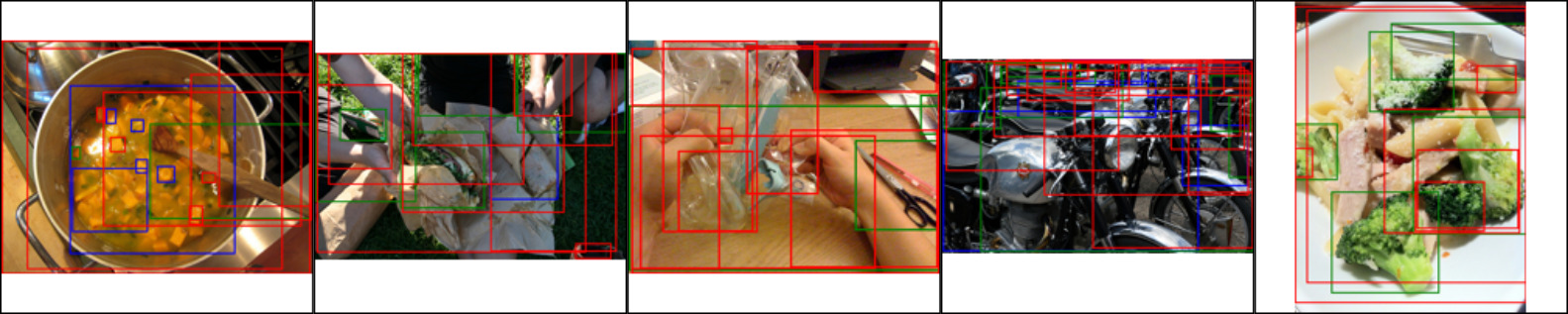} & \raisebox{0.6cm}{$\cdots$} &\includegraphics{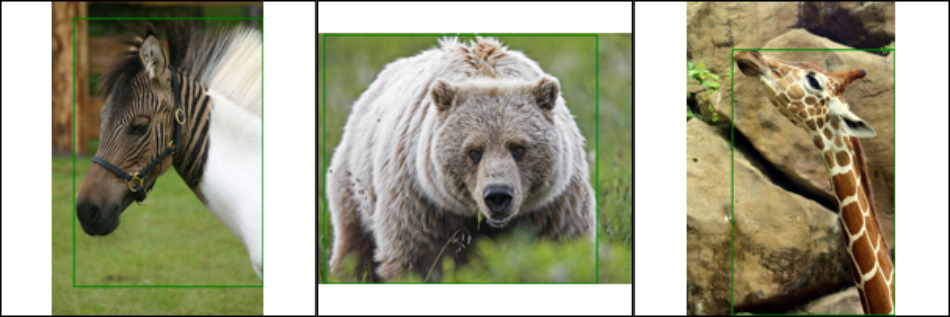} \\
 \cmidrule{2-4} 
 & \multicolumn{3}{c}{\textcolor{red}{\textbf{Query-agnostic Baselines}}} \\
 \cmidrule{2-4} 
 \raisebox{0.6cm}{\textbf{Entropy}} & \includegraphics{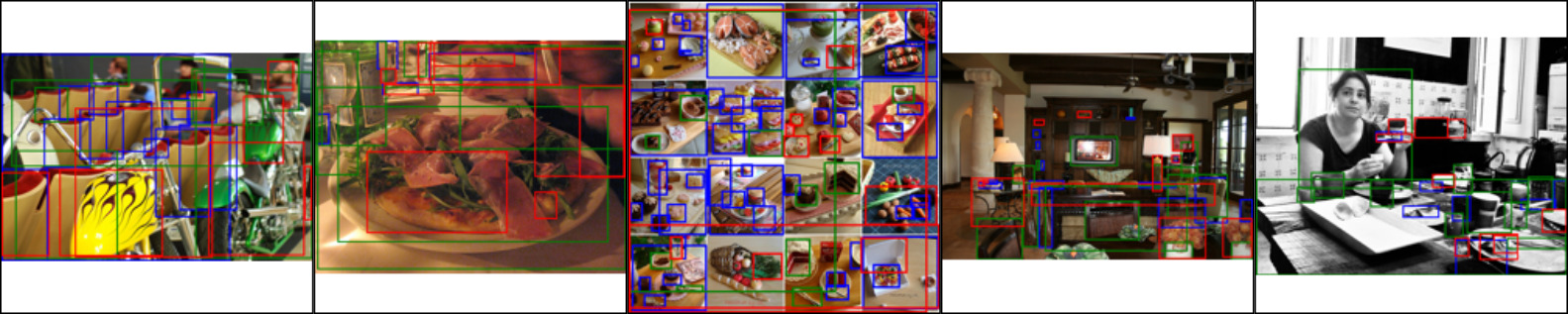} & \raisebox{0.6cm}{$\cdots$} &\includegraphics{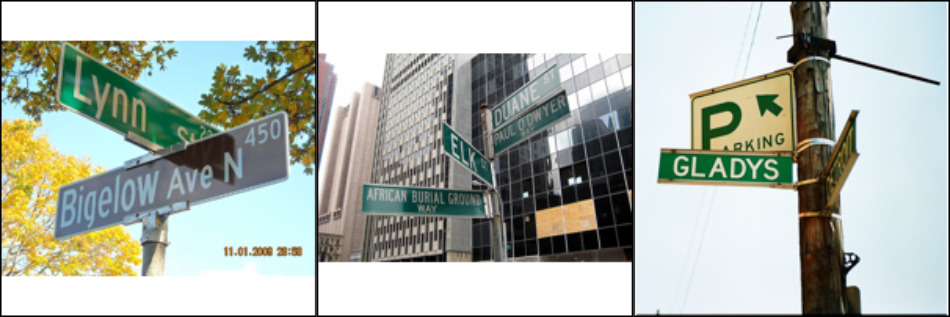} \\
 \raisebox{0.6cm}{\textbf{Dempster-Shafer}} & \includegraphics{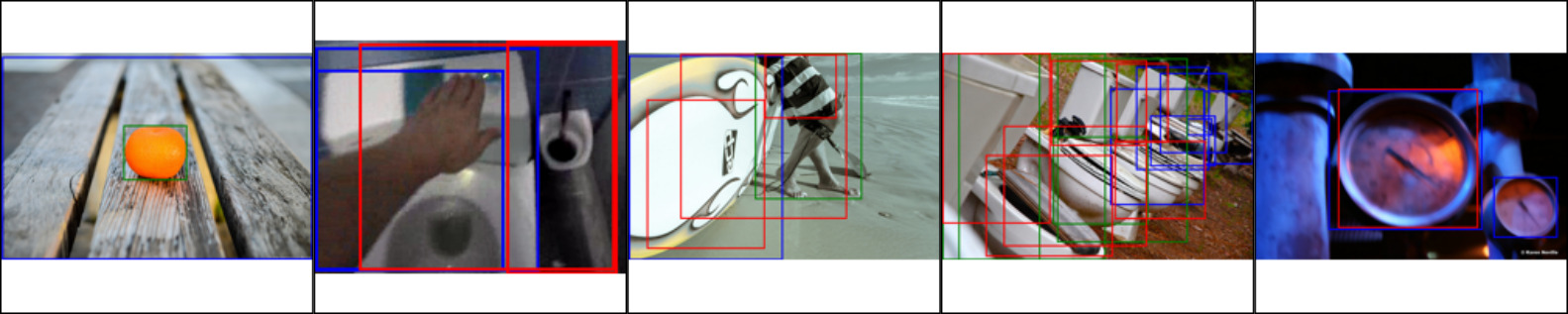} & \raisebox{0.6cm}{$\cdots$} & \includegraphics{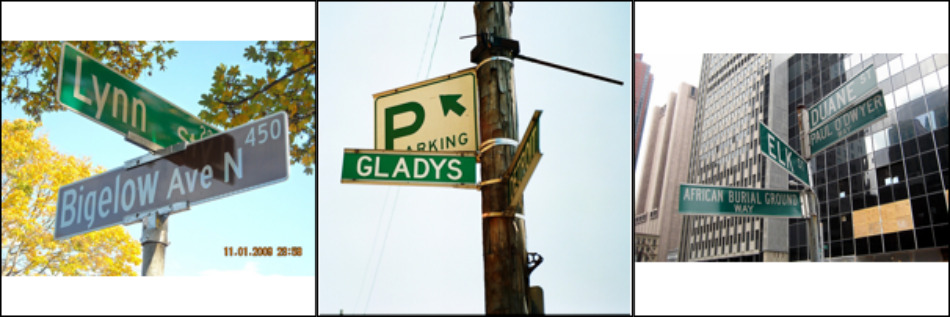} \\
\bottomrule
\end{tabularx}
\caption{
    \label{fig:hardest_big}
Hardest images (left) and easiest images (right) for the \emph{coco-rcnn} detector on the COCO \emph{val} set for different queries.
Entropy and Dempster-Shafer (DS) measures of uncertainty are unable to adapt to different hardness queries and are hence query agnostic.
Green, red, and blue boxes denote true positives, false positives, false negatives, respectively. Note that the ground-truth ranking is different for every query. 
See also Figure~12 in the appendix for examples on the nuImages dataset.}
\end{figure*}

We first study qualitatively the efficacy of our method by searching for the 5 hardest and 3 easiest images for a variety of hardness queries for \emph{coco-rcnn} on the COCO \emph{val 2017} set, and then compare these with the actual hardest images obtained using the ground-truth, as well as those found by the baseline methods.
These results are presented for the error category $\fp$ in Figure~\ref{fig:hardest_big}.
We display for each image the set of true positive, false positive, and false negative bounding boxes in order to allow for visual inspection of the performance of the detector.
We show the corresponding figure for \emph{mmdet-maskrcnn} on nuImages in the appendix.

Contrary to the entropy and DS baselines, our method is able to successfully identify images that have the expected error characteristics for a given query.
For example, when the hardness query is $\texttt{Total}(\texttt{fp})$, which asks for images that are likely to have a large number of false positives, our method finds images featuring a large number of objects (such as books on a shelf, parked cars, etc) many of them being false positives.
This is qualitatively similar to the images ranked using the ground-truth. Interestingly, annotations in the COCO dataset often show shortcomings where objects within a large group are not annotated consistently but we nevertheless observe that our method is able to find such images successfully.

The same is true for the $\texttt{PixelAdj}(\texttt{fp})$ and $\texttt{OccAware}(\texttt{fp})$ queries, where our method finds images that have large false-positive detections and many overlapping boxes, respectively.
Note that in the case of $\texttt{OccAware}(\texttt{fp})$, our method guessed the top two images correctly.
The images identified by the non query-based baselines are not similar to any specific query and mainly seem to feature a large number of bounding boxes in the image, which is easily explained since these methods measure the total uncertainty over all boxes, see Eq.~(\ref{eq:entropy}--\ref{eq:DS}).

The images ranked last in terms of hardness (easiest ones) are somewhat less interesting as they all feature a low number of correctly classified boxes.
It is worth noting that in the case of entropy and DS uncertainty, the lowest ranking images have \emph{no} ground-truth bounding boxes and get a score of zero.
For our method, we show in the appendix that the variance of the scores is sufficient to generate varied pseudo ground-truth configurations, see Fig.~14.

In the Appendix, we also show a histogram (Figure~13) for the estimated and ground-truth false positive bounding-boxes using \emph{coco-rcnn}, and note that, in fact, many images have a hardness score of zero as they are easy for the detector to perform inference on (e.g., single centred object images). Therefore, it becomes crucial to provide a better ranking as there is a small sub-population of potential hard images. Our qualitative experiments below are precisely to capture how well different approaches do in ranking images based on query-based hardness measures. 

\subsection{Query-Based Hard-Image Ranking}
We now evaluate quantitatively the performance of our method at ranking images in order of hardness for a given query.
For any given method, we sort all images in the dataset in decreasing order of estimated hardness scores $\hat q_i$ and compare the resulting ranking to the ground-truth ranking that is obtained by sorting images in decreasing order of ground-truth hardness $q_i$.
We evaluate ranking quality using the normalised discounted cumulative gain (nDCG) which is a well-known metric for ranking-based tasks, see for example~\cite{jarvelin2002cumulated}.
The discounted cumulative gain is defined as follows:
\begin{equation}
    \text{DCG} = \sum_{i=1}^D \frac{\langle q_j \rangle_{j, \hat q_j = \hat q_i}}{\log{_2 (\mathrm{rank}(\bfx_i; \dataset)+1)}},
\end{equation}
where $\mathrm{rank}(\bfx_i; \dataset)$ is the rank of image $i$ when the dataset $\dataset$ is sorted in decreasing order of estimated hardness score $\hat q_i$, and $\langle q_j \rangle_{j, \hat q_j = \hat q_i}$ is the average ground-truth hardness score of all images $\bfx_j$ that have the same estimated hardness score as $\bfx_i$, i.e.~for which $\hat q_j = \hat q_i$.
This ensures that the DCG is independent under re-orderings of images with the same estimated hardness score. 
The normalised DCG is then defined by dividing the gain by the idealised gain as $\text{nDCG} = \text{DCG} / \text{DCG}_{gt}$, where $\text{DCG}_{gt}$ is the DCG of the ground-truth ranking, see~\cite{jarvelin2002cumulated}.

\begin{figure}
    \centering
    \includegraphics[width=\linewidth]{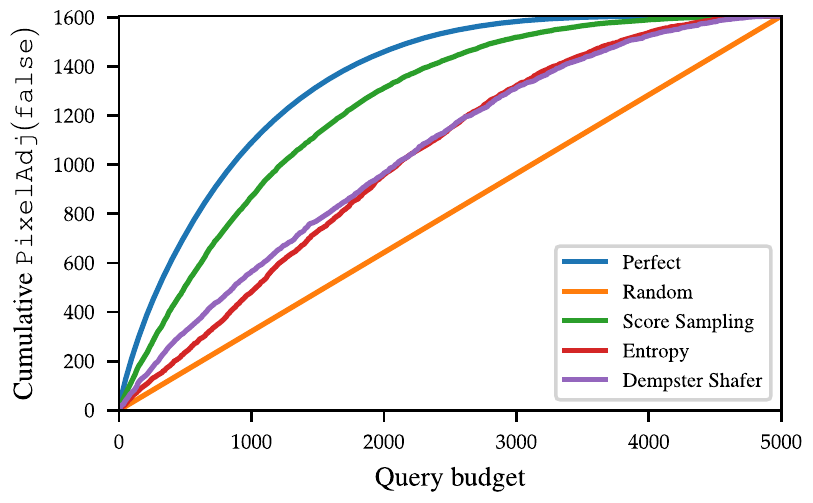}
    \caption{
    Cumulative ground-truth hardness score of images in the COCO \emph{val} set and \emph{coco-rcnn} detector  for images in descending order of estimated hardness scores. $\texttt{PixelAdj}(\false)$ hardness query is used here.
    \label{fig:regret}}
\end{figure}

\begin{table}[t]
\centering
\begin{tabular}{@{}l@{\hspace{4pt}}l@{\hspace{4pt}}l@{\hspace{4pt}}l@{\hspace{4pt}}}
\toprule
     Hardness query &   Entropy &  DS  & SS \\
      &      &     & \textbf{(our)} \\
    \midrule
  &  \multicolumn{3}{c}{Normalised DCG}   \\
  \cmidrule{2-4} 
   & \multicolumn{3}{c}{coco-retina} \\
 \cmidrule{2-4} 
$\texttt{Total}(\texttt{fp})$ & 0.81 & 0.81 & \bf 0.88 \\
$\texttt{Total}(\texttt{fn})$ & 0.91 & \bf 0.92 & \bf 0.92 \\
$\texttt{Total}(\texttt{false})$ & 0.92 & 0.92 & \bf 0.93 \\
$\texttt{PixelAdj}(\texttt{fp})$ & 0.66 & 0.66 & \bf 0.86 \\
$\texttt{PixelAdj}(\texttt{fn})$ & 0.74 & 0.74 & \bf 0.85 \\
$\texttt{PixelAdj}(\texttt{false})$ & 0.76 & 0.76 & \bf 0.87 \\
$\texttt{OccAware}(\texttt{fp})$ & 0.75 & 0.75 & \bf 0.91 \\
$\texttt{OccAware}(\texttt{fn})$ & 0.81 & 0.81 & \bf 0.89 \\
$\texttt{OccAware}(\texttt{false})$ & 0.82 & 0.83 & \bf 0.92 \\
 \cmidrule{2-4} 
   & \multicolumn{3}{c}{coco-rcnn}\\
 \cmidrule{2-4} 
$\texttt{Total}(\texttt{fp})$ & 0.90 & 0.81 & \bf 0.91 \\
$\texttt{Total}(\texttt{fn})$ & \bf 0.90 & 0.77 & \bf 0.90 \\
$\texttt{Total}(\texttt{false})$ & \bf 0.94 & 0.83 & \bf 0.94 \\
$\texttt{PixelAdj}(\texttt{fp})$ & 0.79 & 0.80 & \bf 0.93 \\
$\texttt{PixelAdj}(\texttt{fn})$ & 0.72 & 0.77 & \bf 0.81 \\
$\texttt{PixelAdj}(\texttt{false})$ & 0.81 & 0.84 & \bf 0.90 \\
$\texttt{OccAware}(\texttt{fp})$ & 0.85 & 0.77 & \bf 0.94 \\
$\texttt{OccAware}(\texttt{fn})$ & 0.83 & 0.70 & \bf 0.86 \\
$\texttt{OccAware}(\texttt{false})$ & 0.88 & 0.78 & \bf 0.93 \\
\bottomrule
\end{tabular}
 \begin{tabular}{@{}l@{\hspace{4pt}}l@{\hspace{4pt}}l@{\hspace{4pt}}}
\toprule
        Entropy &  DS  & SS \\
            &     & \textbf{(our)} \\
    \midrule
    \multicolumn{3}{c}{Normalised DCG}   \\
  \cmidrule{1-3} 
   \multicolumn{3}{c}{mmdet-maskrcnn}\\
 \cmidrule{1-3} 
0.92 & 0.88 & \bf 0.95 \\
\bf 0.89 & 0.83 & \bf 0.89 \\
0.95 & 0.89 & \bf 0.96 \\
0.52 & 0.55 & \bf 0.86 \\
0.50 & 0.51 & \bf 0.75 \\
0.58 & 0.61 & \bf 0.82 \\
0.87 & 0.83 & \bf 0.96 \\
0.81 & 0.74 & \bf 0.85 \\
0.89 & 0.83 & \bf 0.94 \\
 \cmidrule{1-3} 
   \multicolumn{3}{c}{mmdet-cascade}\\
 \cmidrule{1-3} 
0.92 & 0.87 & \bf 0.94 \\
\bf 0.89 & 0.83 & \bf 0.89 \\
\bf 0.95 & 0.89 & \bf 0.95 \\
0.46 & 0.49 & \bf 0.83 \\
0.51 & 0.52 & \bf 0.75 \\
0.56 & 0.58 & \bf 0.84 \\
0.84 & 0.80 & \bf 0.94 \\
0.81 & 0.75 & \bf 0.86 \\
0.87 & 0.81 & \bf 0.93 \\
\bottomrule
\end{tabular}

\caption{
Ranking performance of Entropy, Dempster-Shafer, and Score sampling (Our) measured by the Normalised Discounted Cumulative Gain (nDCG) against the ground-truth ranking for different $\texttt{Total}$, $\texttt{PixelAdj}$, and $\texttt{OccAware}$ queries.
Higher is better.
}
\label{tab:continous_table}
\end{table}

In Table~\ref{tab:continous_table}, we observe that in practically all the cases our method (SS) matches or exceeds (significantly) the performance of the entropy and the DS baselines.
Since the baseline methods do not allow query-based hardness computation, we do not expect these techniques to perform well on the pixel-aware and occlusion-adjusted hardness queries. This is evident in Table~\ref{tab:continous_table} for the pixel-weighted and overlap-adjusted measures of hardness.
The gap between methods is much smaller in the case of the $\texttt{Total}$ queries which indicates that the entropy and DS are a good proxy for finding the total number of errors in an image. In particular, entropy outperforms our method slightly for the query $\texttt{Total}(\texttt{fn})$ which could be due to the fact that the detectors are not perfectly calibrated.

\paragraph{Ranking performance using query-budget} Another way to visualise the ranking performance would be to consider the cumulative ground-truth hardness of all the images mined within a certain query budget. This allows us to easily visualise \textit{how quickly} hard images can be found within a dataset by each method.
We show this for the $\texttt{Pixel}(\texttt{false})$ query and the \emph{coco-rcnn} detector in Figure~\ref{fig:regret} and observe that our method not just provides better ranking (as shown in Table~\ref{tab:continous_table}), but also provides the hard images much quicker than the other baselines. We note that the entropy and the DS baselines perform similarly, and are barely better than a random ranking, which corresponds to the diagonal line through the origin.
We display similar figures for other detectors, datasets and hardness definitions in the appendix (Figures~6--9).

\subsection{Query-Based Hard-Image Classification}

We finally consider the task of \textit{finding} hard images for a specific detector and hardness query within a dataset.
Given a definition of hardness, an image is defined to be ``hard'' if its ground-truth hardness $q_i$ is above a chosen threshold $t_{\text{hard}}$, and ``easy'' otherwise.
This effectively assigns a binary class label to each image and we evaluate the performance of our method at solving this \emph{hard vs.~easy} classification task.
To do so, we construct a binary classifier by introducing a threshold $t_{\text{score}}$ on the estimated hardness score, and classify an image as ``hard'' if $\hat q_i > t_{\text{score}}$ and ``easy'' otherwise.
Performance of this binary classifier can be evaluated using typical metrics for binary classification, such as AUROC, across all possible score thresholds $t_{\text{score}}$.
If needed, a score threshold with a given application-specific trade-off of precision and recall can be easily obtained using a held-out dataset.

We evaluate our method and baselines across different hard vs.~easy image ratios by choosing the hardness threshold $t_{\text{hard}}$ such that this ratio is $5\%$, $10\%$, $25\%$, and $50\%$ for each dataset.
We present the mean area under the ROC curve (mAUROC) over these threshold choices in Table~2 in the appendix.
In almost all cases, score sampling exceeds or matches the performance of the baseline techniques similarly as for our previous ranking experiments.
Again, we observe score sampling performs much better for hardness definitions which measure a specific property of the detection error.

\section{Conclusive Remarks}
In this work, we proposed a framework to enable finding query-based hard images for object detectors without having access to ground-truth labels. We showed that it is possible to construct a distribution over pseudo ground-truth labels and quantify a given query-based hardness measure via an efficient Monte Carlo approximation.  We provided extensive analysis to show such approximations rank and find hard images more effectively than general purpose image hardness estimation techniques such as entropy and Dempster-Shafer. 

We hope that our work opens possibilities to explore situations where query-based hardness measures are used in practice to improve downstream tasks and also to understand their impact on data-shift scenarios. One crucial assumption in our approach is that the object detectors are considered to be well enough calibrated. Therefore, another future work entails exploring differently calibrated models and understanding their impact on the hardness measures as well as studying the efficacy of this technique on out of domain data.
Finally, a more complex and accurate stochastic model could also be devised to provide improved ranking performance.  


\section*{Acknowledgements}
We gratefully thank Sina Samangooei and Viveka Kulharia for valuable discussions at the inception of this project. 
We also wish to thank Kemal Oksuz, Nicholas A.~Lord, Anuj Sharma, and Zygmunt Lenyk for their valuable time to provide comments on this document. 

\bibliography{bibliography}

\FloatBarrier
\clearpage
\include{appendix}

\end{document}

%% file: appendix.tex
\appendix
\onecolumn
\section*{Appendix}

\section{Query-Based Hard-Image Classification Results}

Table~\ref{tab:binaryroc} shows the mean area under the ROC curve (mAUROC) over multiple hardness threshold as discussed in the main experiments section.

\begin{table}
\centering
\begin{tabular}{@{}l@{\hspace{4pt}}l@{\hspace{4pt}}l@{\hspace{4pt}}l@{\hspace{4pt}}l@{\hspace{4pt}}}
\toprule
         Hardness query & Entropy & DS & SS  \\
       &   &  & (our)   \\
    \midrule
     &\multicolumn{3}{c}{mAUROC} \\
    \cmidrule{2-4}
   &\multicolumn{3}{c}{coco-retina } \\
 \cmidrule{2-4}
$\texttt{Total}(\texttt{fp})$ & 0.90 & 0.73 & \bf 0.94 \\
$\texttt{Total}(\texttt{fn})$ & 0.89 & 0.75 & \bf 0.90 \\
$\texttt{Total}(\texttt{false})$ & 0.93 & 0.75 & \bf 0.95 \\
$\texttt{PixelAdj}(\texttt{fp})$ & 0.67 & 0.68 & \bf 0.93 \\
$\texttt{PixelAdj}(\texttt{fn})$ & 0.73 & 0.75 & \bf 0.82 \\
$\texttt{PixelAdj}(\texttt{false})$ & 0.70 & 0.72 & \bf 0.90 \\
$\texttt{OccAware}(\texttt{fp})$ & 0.84 & 0.70 & \bf 0.95 \\
$\texttt{OccAware}(\texttt{fn})$ & 0.87 & 0.72 & \bf 0.89 \\
$\texttt{OccAware}(\texttt{false})$ & 0.88 & 0.72 & \bf 0.96 \\
 \cmidrule{2-4}
   &\multicolumn{3}{c}{coco-rcnn}\\
 \cmidrule{2-4}
$\texttt{Total}(\texttt{fp})$ & 0.90 & 0.73 & \bf 0.94 \\
$\texttt{Total}(\texttt{fn})$ & 0.89 & 0.75 & \bf 0.90 \\
$\texttt{Total}(\texttt{false})$ & 0.93 & 0.75 & \bf 0.95 \\
$\texttt{PixelAdj}(\texttt{fp})$ & 0.67 & 0.68 & \bf 0.93 \\
$\texttt{PixelAdj}(\texttt{fn})$ & 0.73 & 0.75 & \bf 0.82 \\
$\texttt{PixelAdj}(\texttt{false})$ & 0.70 & 0.72 & \bf 0.90 \\
$\texttt{OccAware}(\texttt{fp})$ & 0.84 & 0.70 & \bf 0.95 \\
$\texttt{OccAware}(\texttt{fn})$ & 0.87 & 0.72 & \bf 0.89 \\
$\texttt{OccAware}(\texttt{false})$ & 0.88 & 0.72 & \bf 0.96 \\
   \bottomrule
   \end{tabular}
\begin{tabular}{@{}l@{\hspace{4pt}}l@{\hspace{4pt}}l@{\hspace{4pt}}l@{\hspace{4pt}}}
   \toprule
            Entropy & DS & SS  \\
          &  & (our)   \\
\midrule
     \multicolumn{3}{c}{mAUROC} \\
     \cmidrule{1-3}
   \multicolumn{3}{c}{mmdet-maskrcnn} \\
 \cmidrule{1-3}
0.89 & 0.80 & \bf 0.93 \\
\bf 0.85 & 0.78 & \bf 0.85 \\
0.92 & 0.83 & \bf 0.93 \\
0.76 & 0.74 & \bf 0.90 \\
0.74 & 0.71 & \bf 0.78 \\
0.75 & 0.74 & \bf 0.88 \\
0.86 & 0.81 & \bf 0.96 \\
0.83 & 0.77 & \bf 0.86 \\
0.88 & 0.81 & \bf 0.93 \\
 \cmidrule{1-3}
   \multicolumn{3}{c}{mmdet-cascade}\\
   \cmidrule{1-3}
0.87 & 0.79 & \bf 0.92 \\
\bf 0.86 & 0.79 & 0.85 \\
0.91 & 0.83 & \bf 0.92 \\
0.75 & 0.72 & \bf 0.87 \\
0.74 & 0.71 & \bf 0.80 \\
0.74 & 0.73 & \bf 0.87 \\
0.85 & 0.80 & \bf 0.95 \\
0.83 & 0.78 & \bf 0.87 \\
0.87 & 0.81 & \bf 0.93 \\
\bottomrule
\end{tabular}
\caption{Mean area under the ROC curve (mAUROC) averaged over different hard-image rates for the binary \emph{easy vs.~hard} classification task described in the main text.
Results are shown for Entropy, Dempster-Shafer, and Score sampling (Our) for different $\texttt{Total}$, $\texttt{PixelAdj}$, and $\texttt{OccAware}$ queries.
Higher is better for all metrics.
}
\label{tab:binaryroc}
\end{table}

\begin{figure}[t]
     \centering
        \subfigure[Calibration of detectors on nuImages.]{\includegraphics[width=0.45\linewidth]{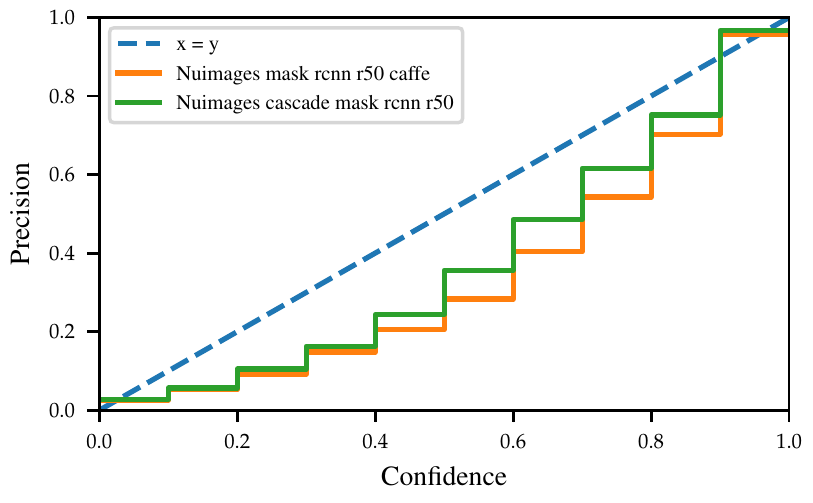}}
     \subfigure[Calibration of detectors on coco.]{\includegraphics[width=0.45\linewidth]{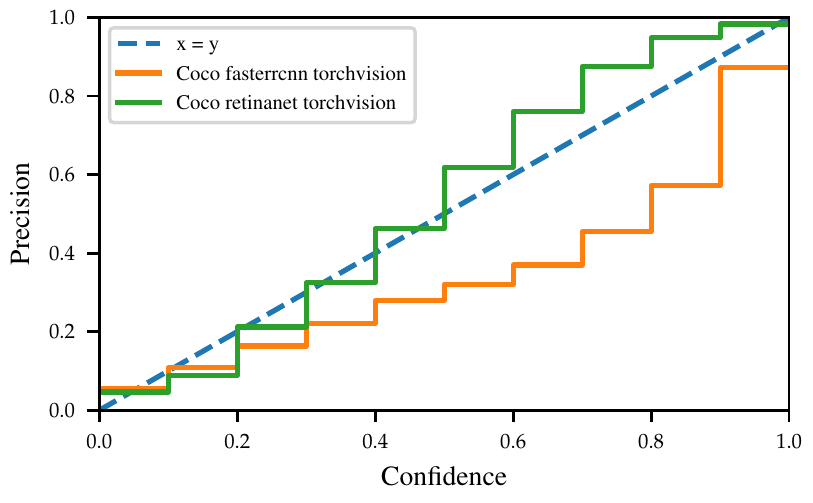}}
        \caption{Confidence histogram plots for pre-trained detectors on test datasets.}
        \label{fig:ece}
\end{figure}

\section{Detector Calibration}
\label{sec:calibration}

Figure~\ref{fig:ece} shows confidence histogram plots for each of the detectors used for the experiments in the evaluation section.
These confidence histograms are produced by taking all of the bounding boxes for each detector-dataset combination and attempting to associate them with ground truth annotations; binning the detected boxes by score enables us to calculate the precision for each bin.
For most bins, the detectors are overconfident in their detections, i.e. the rate of detection is lower than that which would be implied from a probabilistic interpretation of the detected box scores.
However we note that the accuracy monotonically increases as confidence increases, i.e. the \emph{ordering} of the boxes by score is approximately correct.
Although one would expect this miscalibration to have a dire effect on the efficacy of score sampling, because the number of false positives should be underestimated and the number of false negatives overestimated, in practice our experiments show that score sampling is an effective method of identifying hard images.

\section{Cumulative Hardness by Query Budget}
\label{sec:regret}

\begin{figure*}
    \centering
\includegraphics[width=0.30\linewidth]{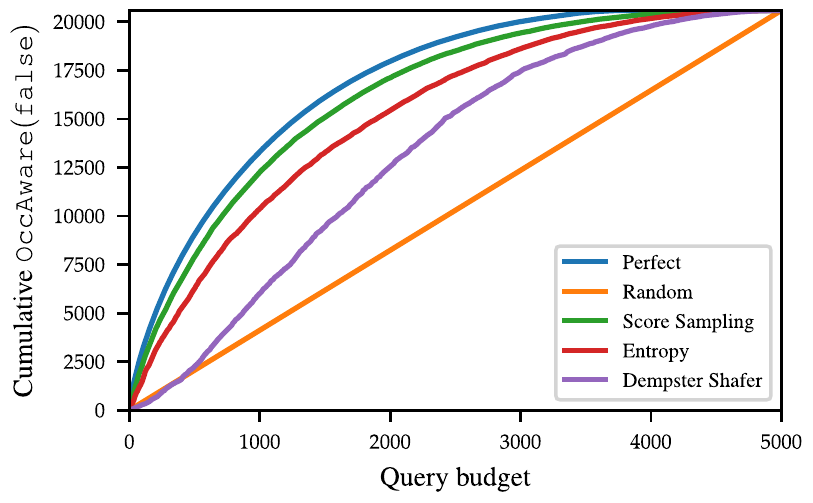}~\includegraphics[width=0.30\linewidth]{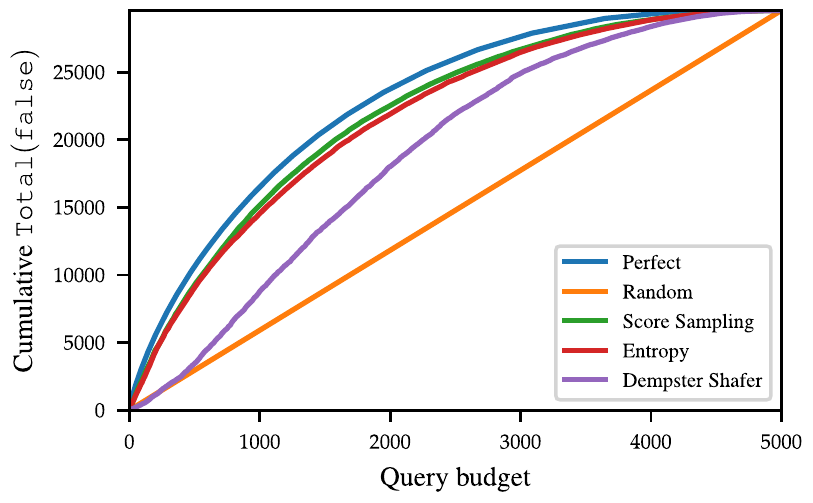}~\includegraphics[width=0.30\linewidth]{results/regret/regret_Coco_fasterrcnn_torchvision_false_pixel.pdf} \\
\includegraphics[width=0.30\linewidth]{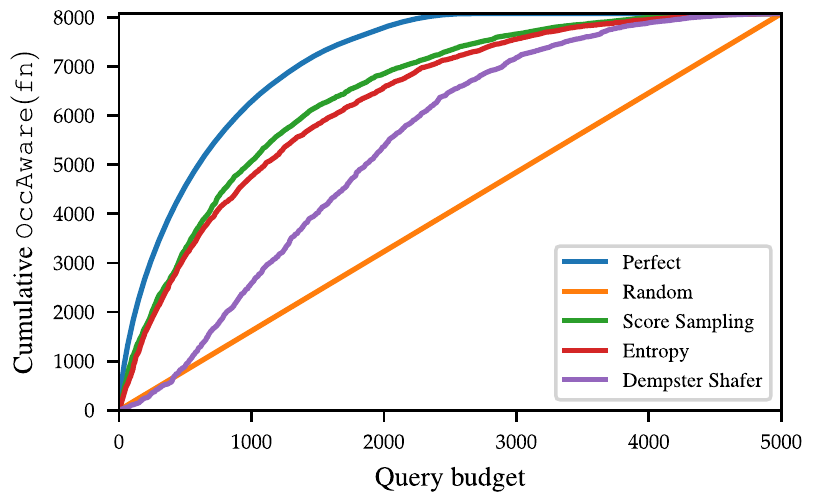}~\includegraphics[width=0.30\linewidth]{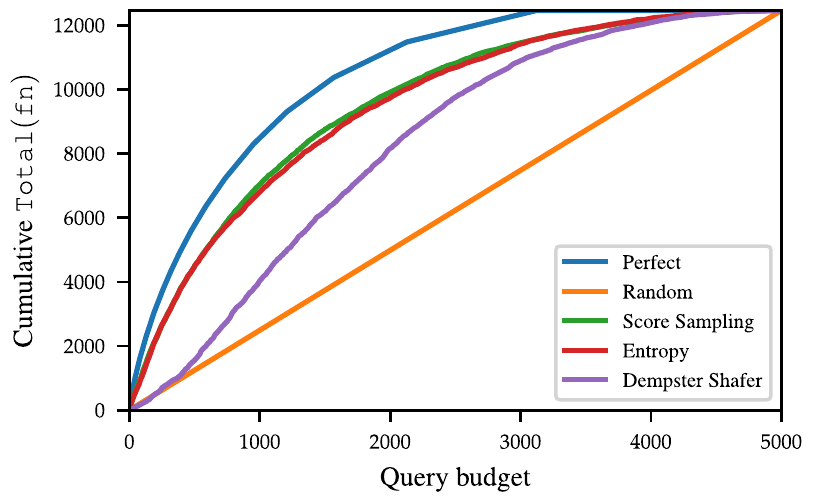}~\includegraphics[width=0.30\linewidth]{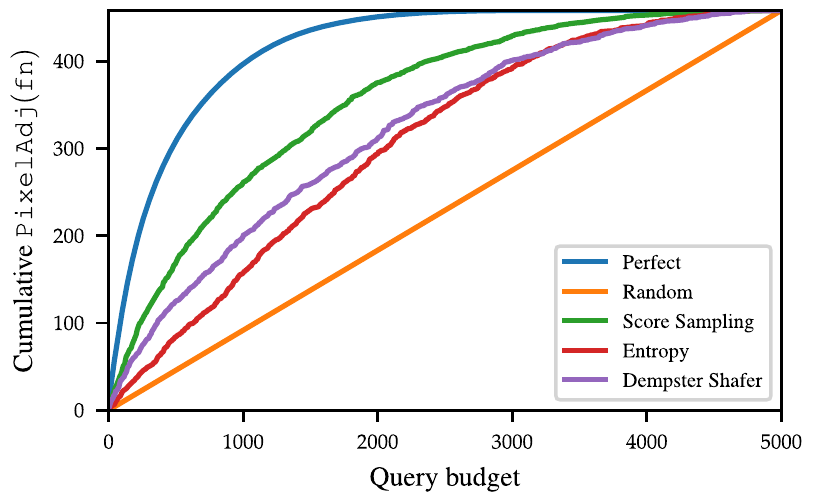}  \\
\includegraphics[width=0.30\linewidth]{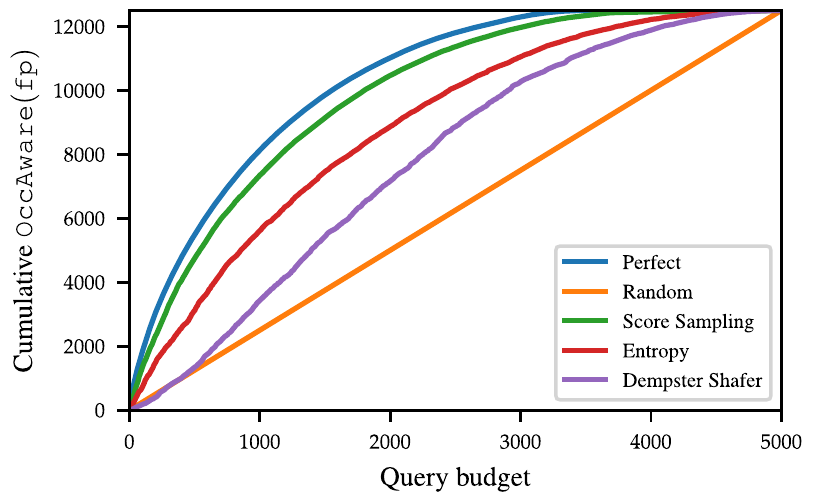}~\includegraphics[width=0.30\linewidth]{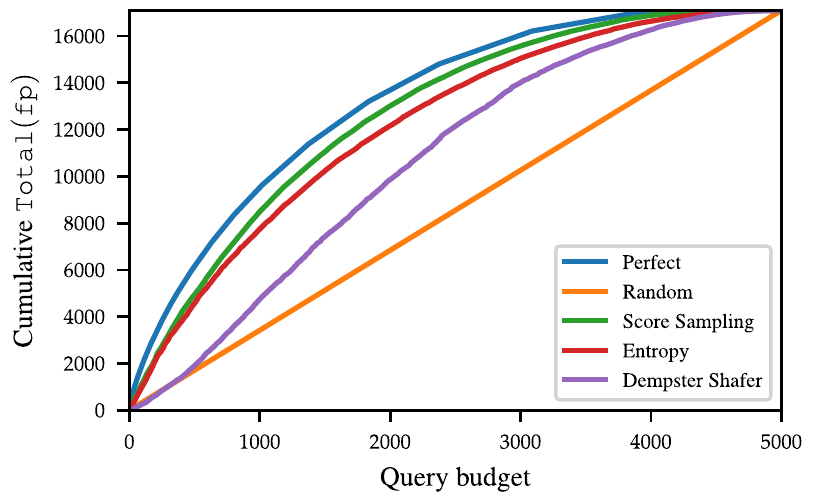}~\includegraphics[width=0.30\linewidth]{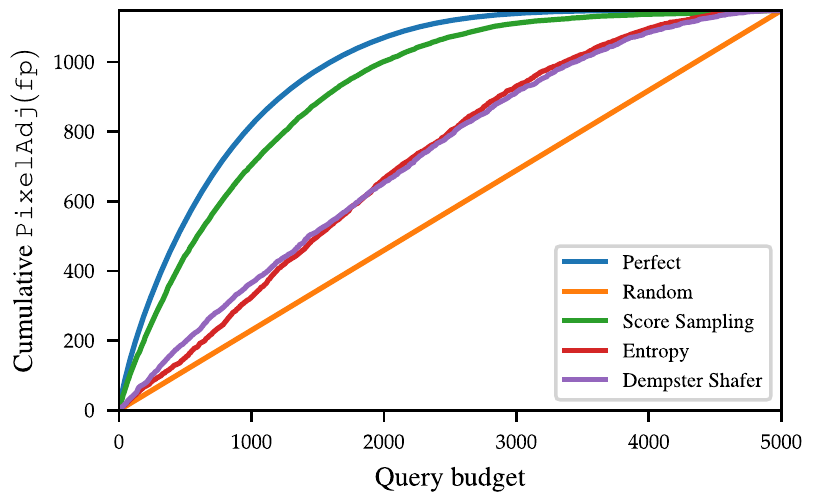} 
    \caption{Cumulative ground-truth hardness of boxes in images obtained with a fixed query budget for the overlap-adjusted hardness definition for \emph{coco-rcnn}.}
    \label{fig:regret_1}
\end{figure*}

\begin{figure*}
    \centering
\includegraphics[width=0.30\linewidth]{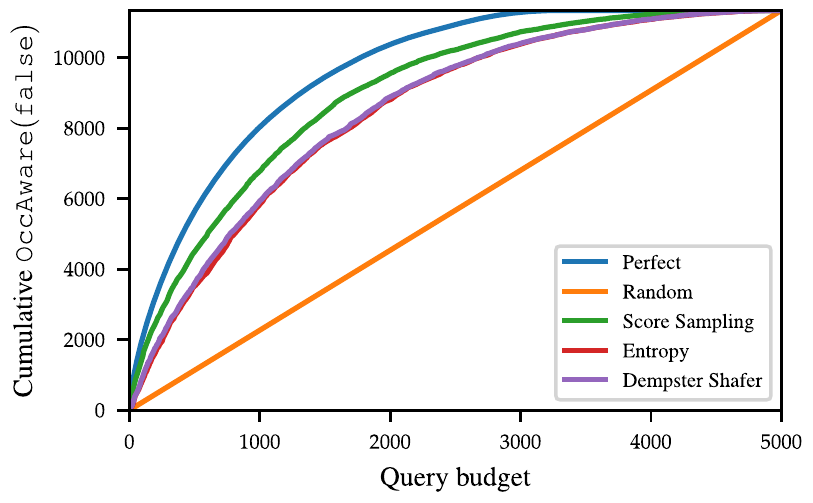}~\includegraphics[width=0.30\linewidth]{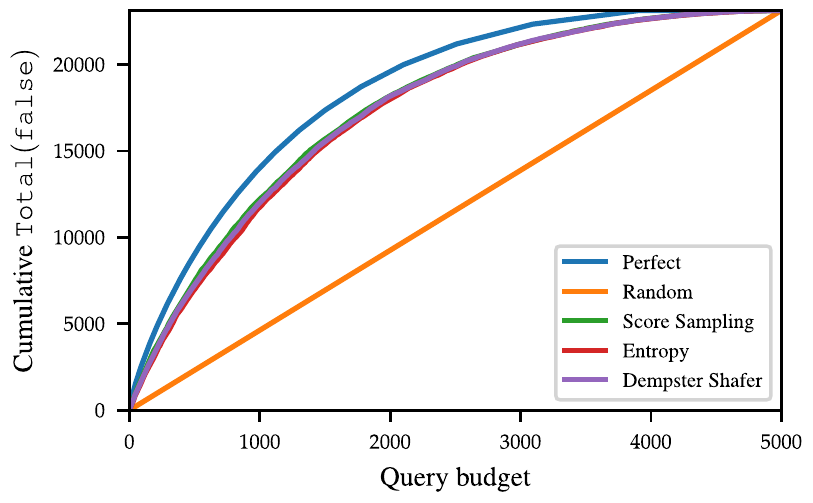}~\includegraphics[width=0.30\linewidth]{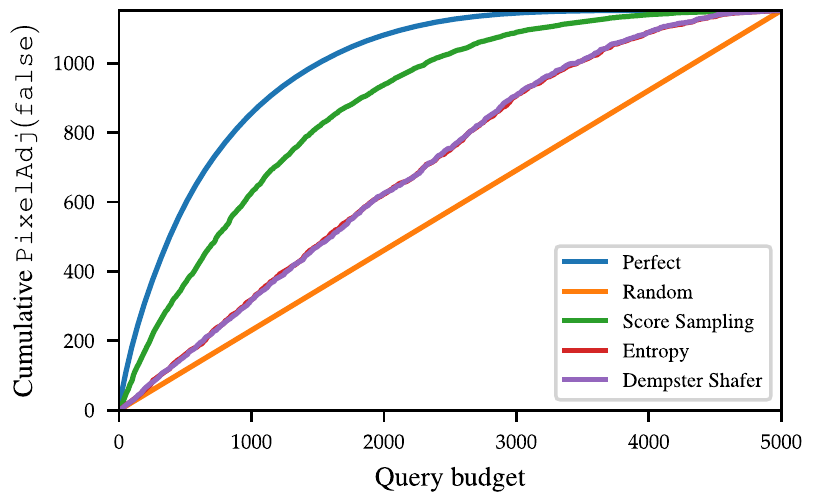}      \\

\includegraphics[width=0.30\linewidth]{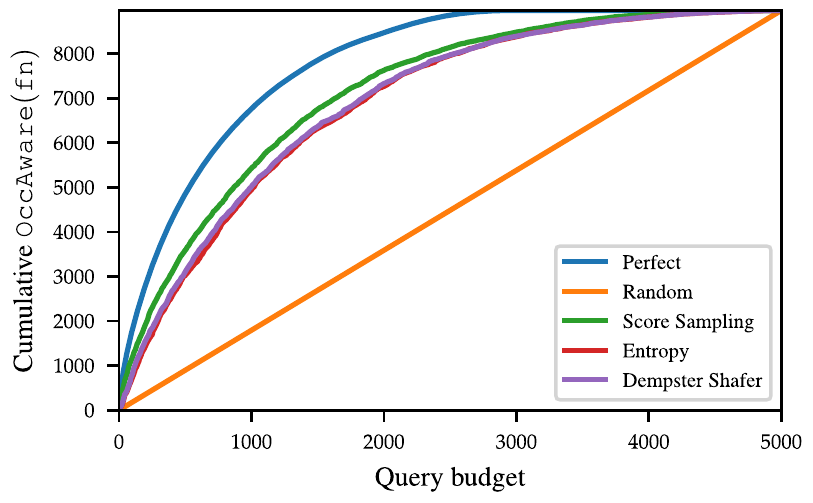}~\includegraphics[width=0.30\linewidth]{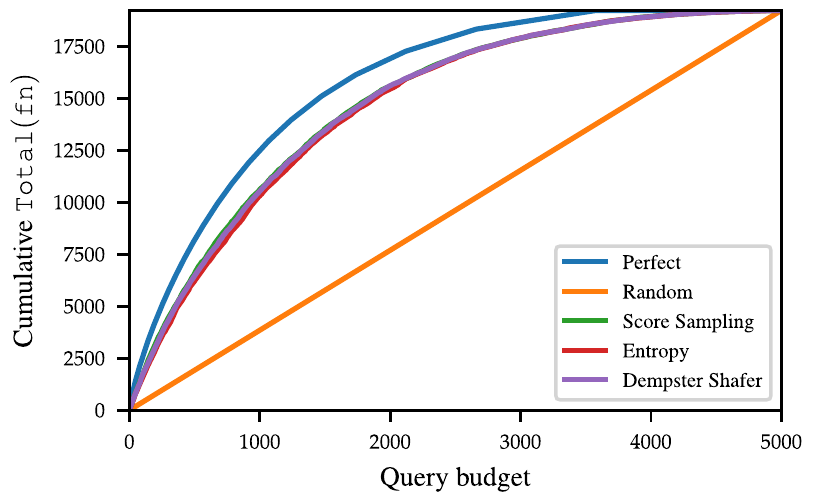}~\includegraphics[width=0.30\linewidth]{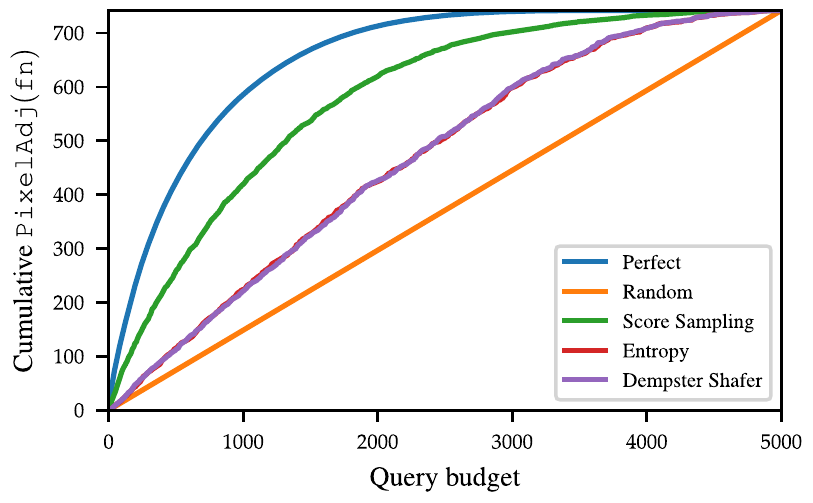}      \\

\includegraphics[width=0.30\linewidth]{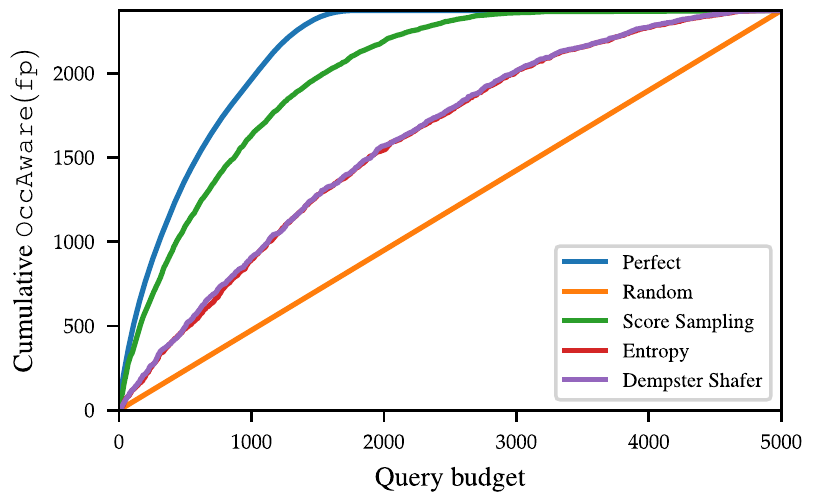}~\includegraphics[width=0.30\linewidth]{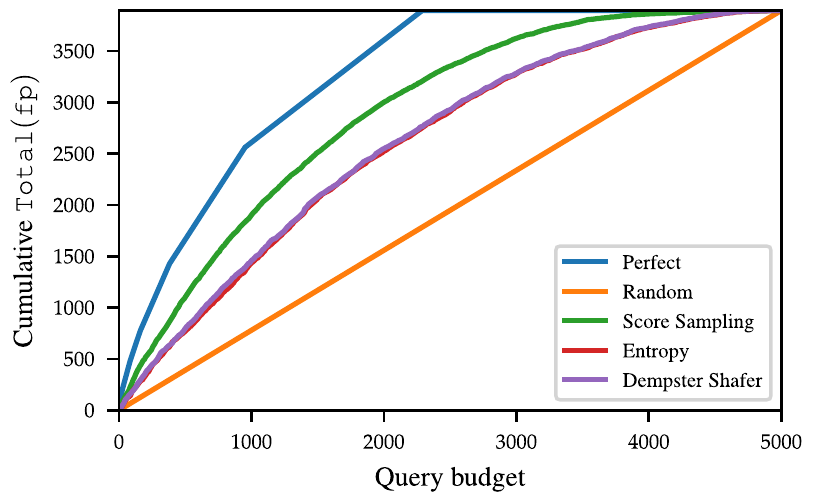}~\includegraphics[width=0.30\linewidth]{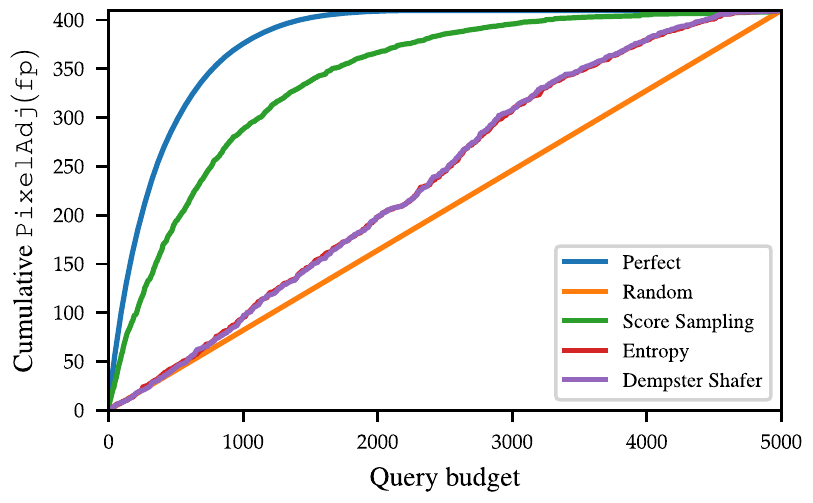}
    \caption{Cumulative ground-truth hardness of boxes in images obtained with a fixed query budget for the overlap-adjusted hardness definition for \emph{coco-retina}.}
   \label{fig:regret_2}
\end{figure*}

\begin{figure*}
    \centering
\includegraphics[width=0.30\linewidth]{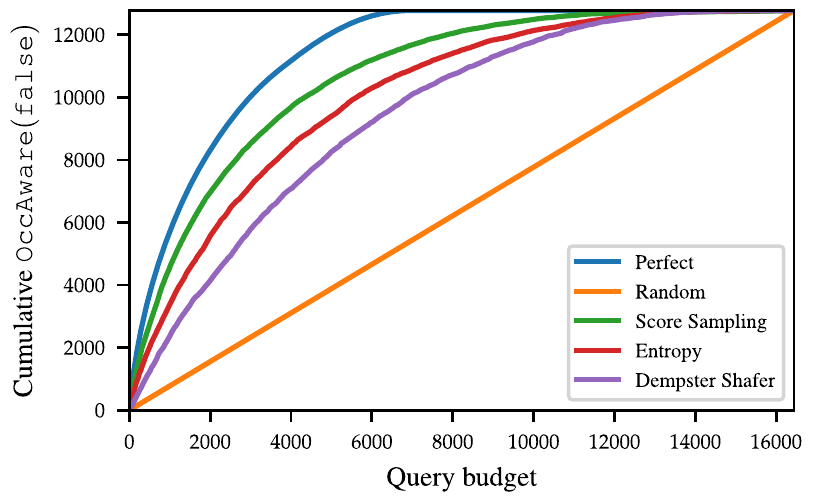}~\includegraphics[width=0.30\linewidth]{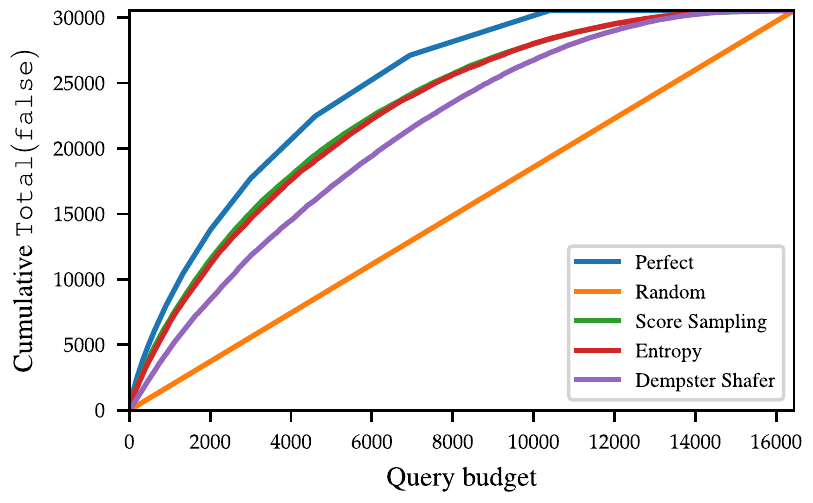}~\includegraphics[width=0.30\linewidth]{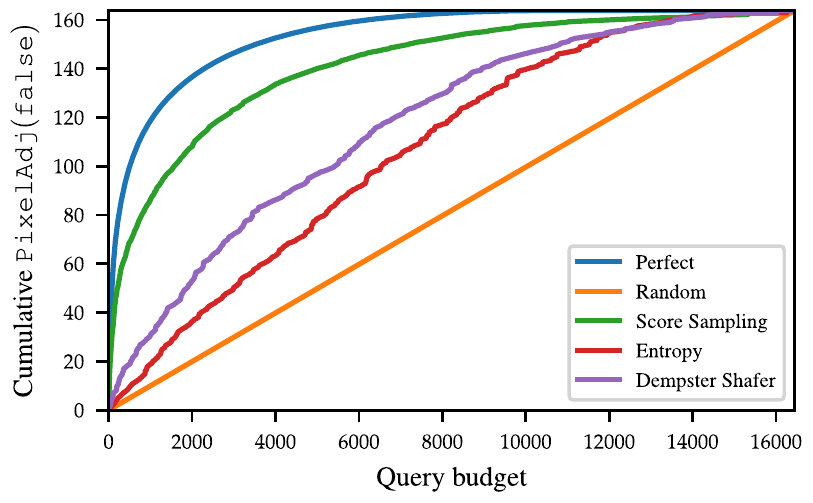} \\

\includegraphics[width=0.30\linewidth]{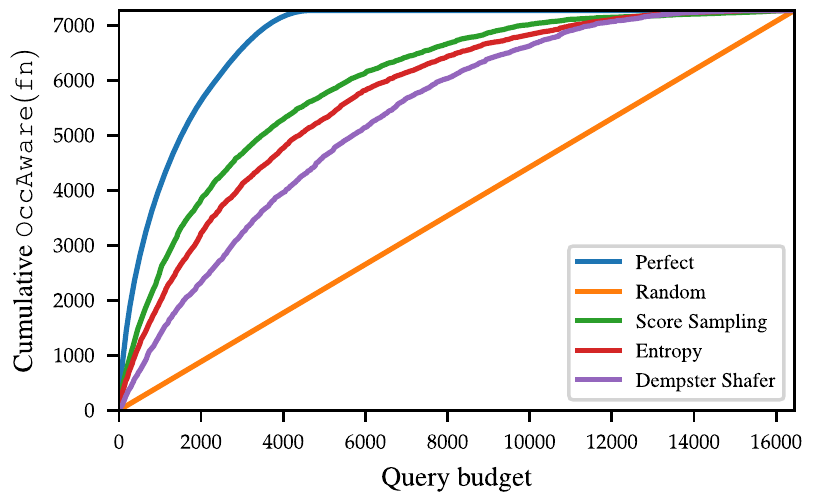}~\includegraphics[width=0.30\linewidth]{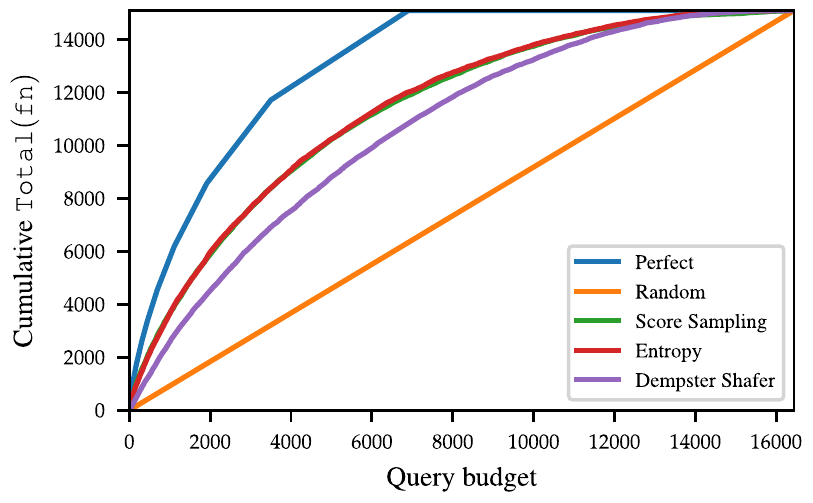}~\includegraphics[width=0.30\linewidth]{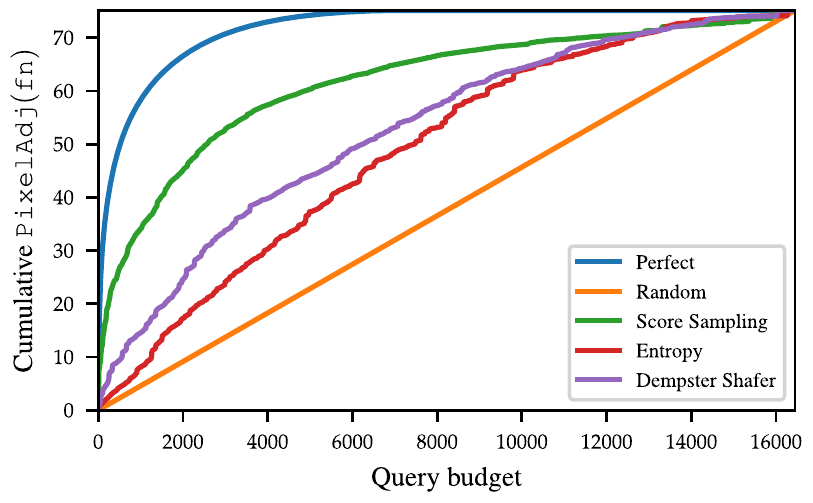}     \\

\includegraphics[width=0.30\linewidth]{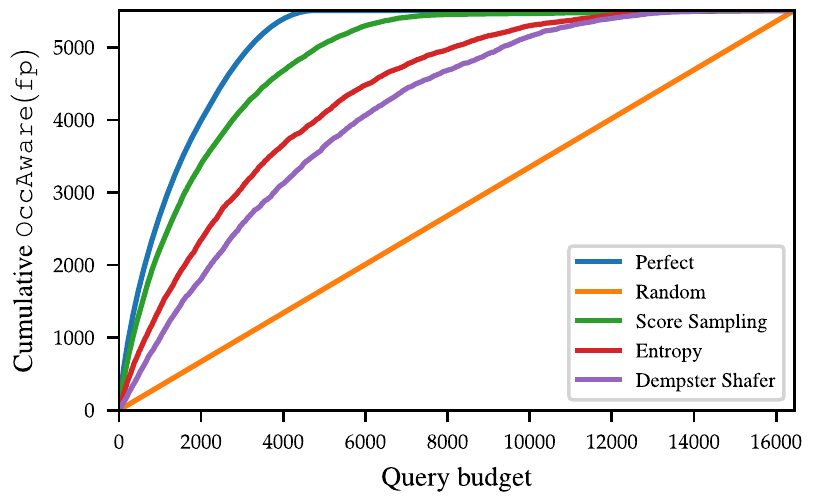}~\includegraphics[width=0.30\linewidth]{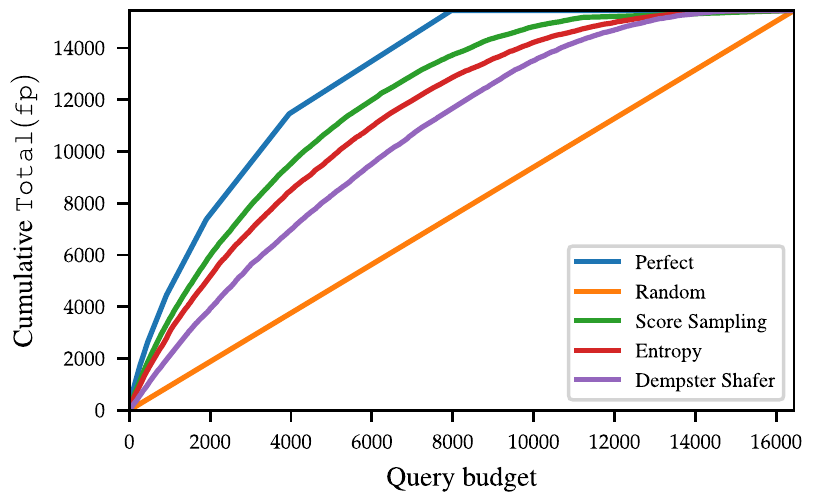}~\includegraphics[width=0.30\linewidth]{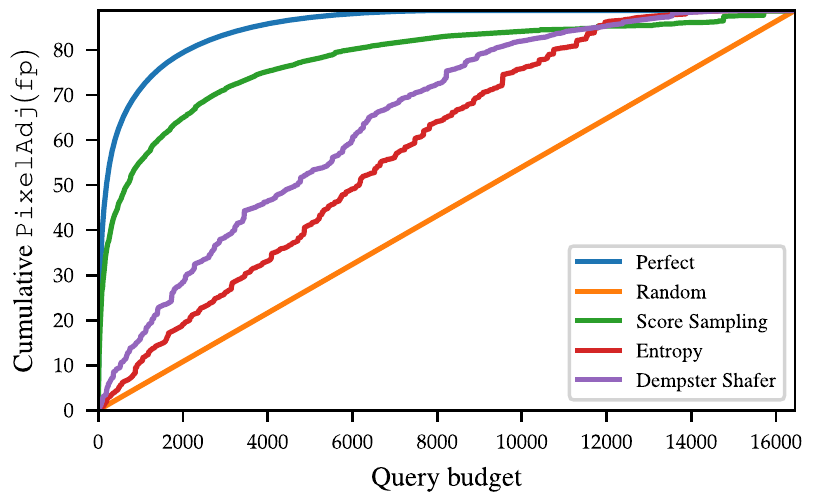} 
    \caption{Cumulative ground-truth hardness of boxes in images obtained with a fixed query budget for the overlap-adjusted hardness definition for \emph{mmdet-cascade}.}
   \label{fig:regret_3}
\end{figure*}

\begin{figure*}
    \centering
\includegraphics[width=0.30\linewidth]{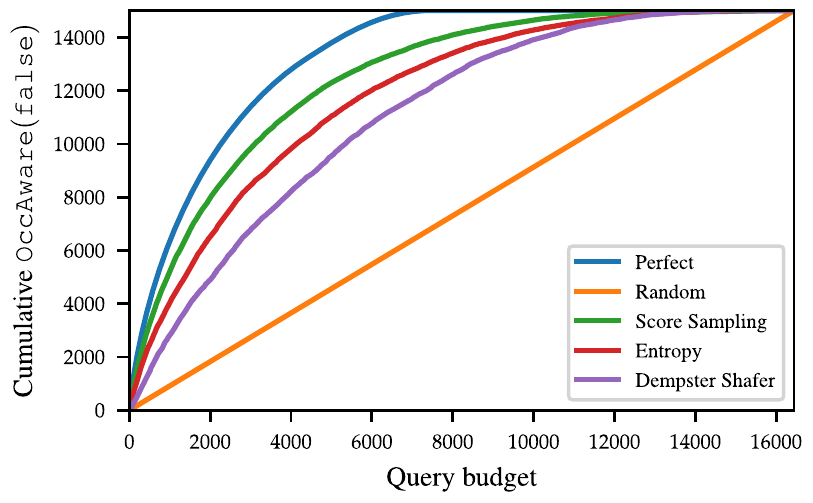}~\includegraphics[width=0.30\linewidth]{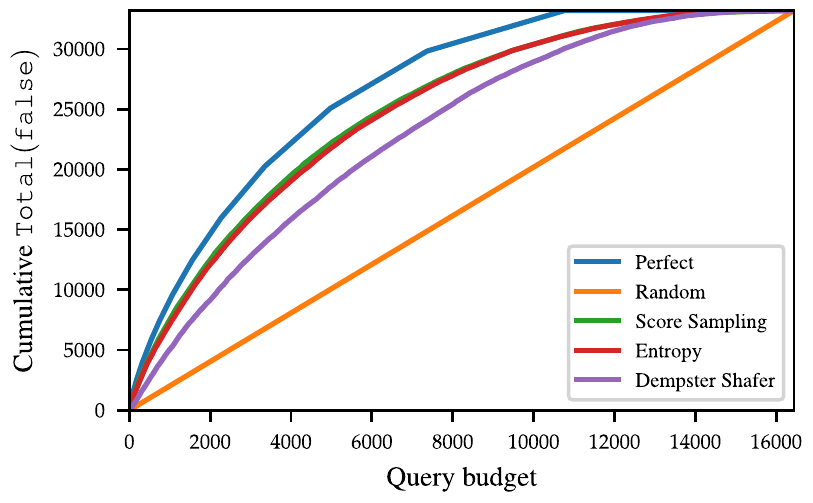}~\includegraphics[width=0.30\linewidth]{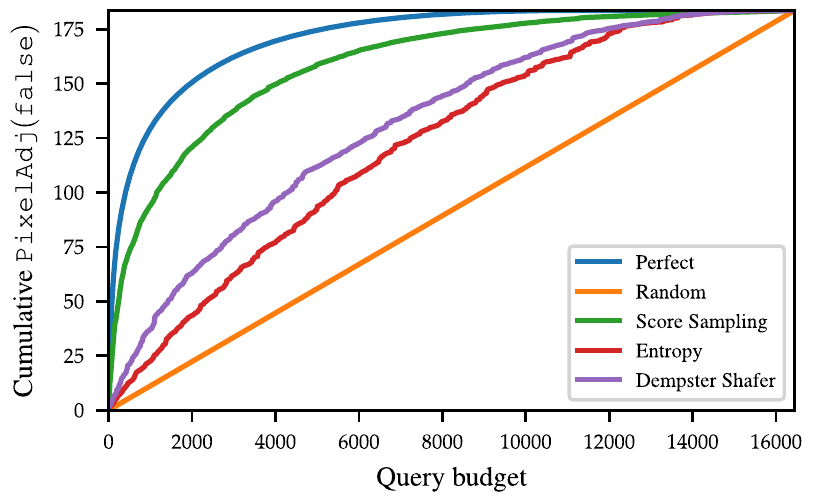}  \\

\includegraphics[width=0.30\linewidth]{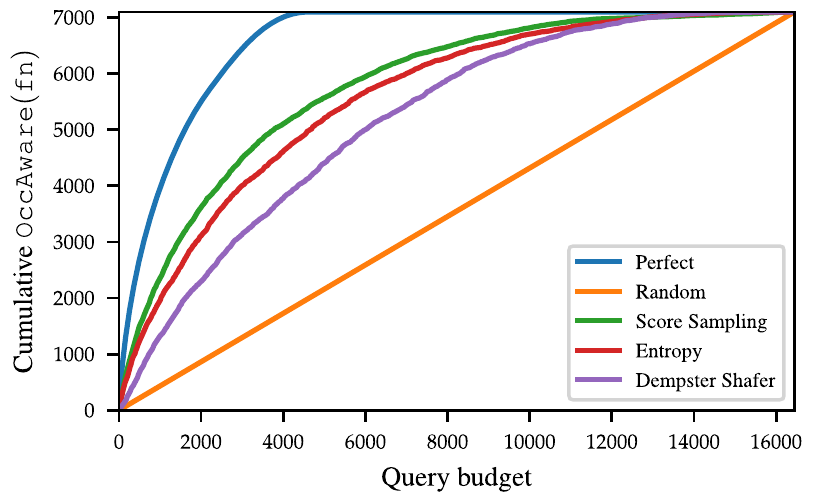}~\includegraphics[width=0.30\linewidth]{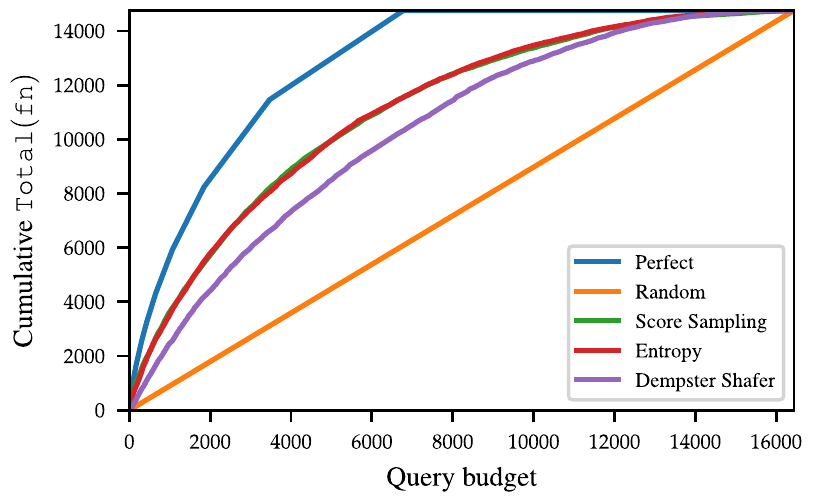}~\includegraphics[width=0.30\linewidth]{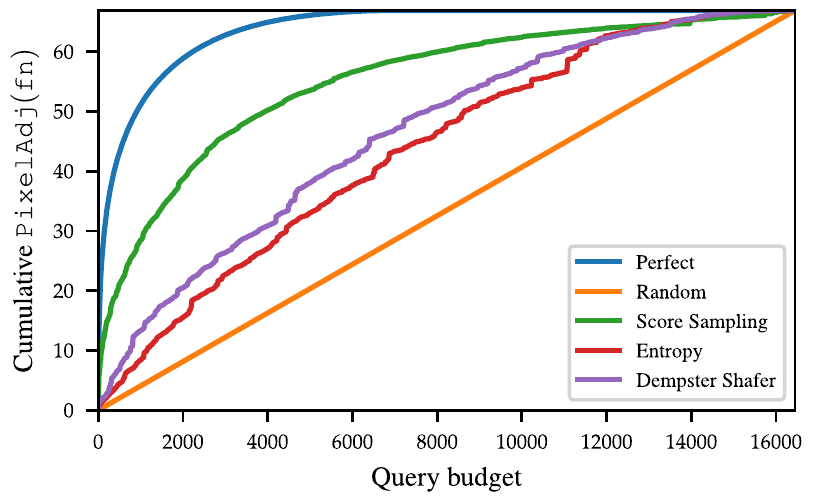}  \\

\includegraphics[width=0.30\linewidth]{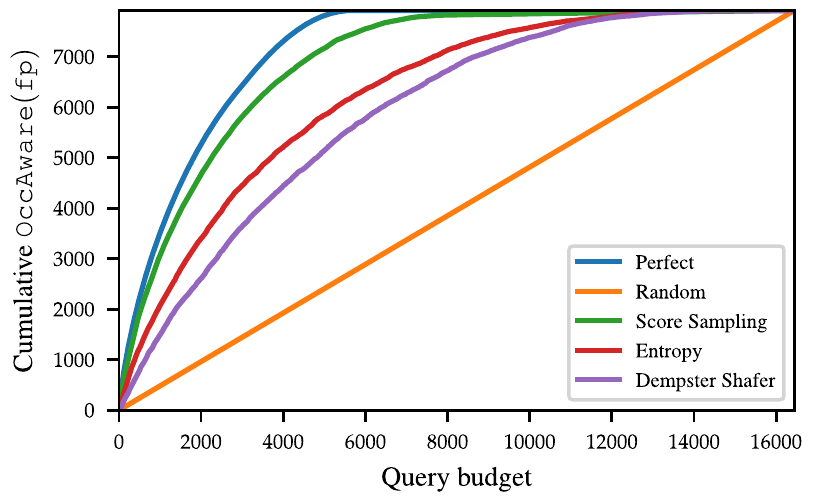}~\includegraphics[width=0.30\linewidth]{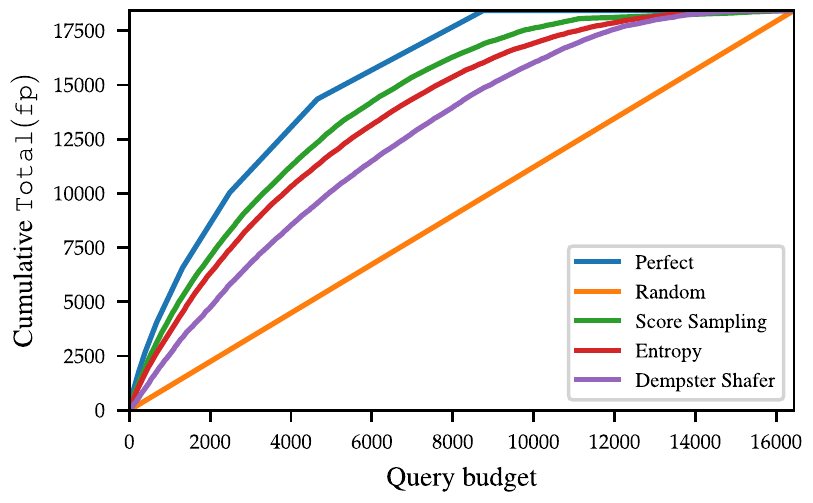}~\includegraphics[width=0.30\linewidth]{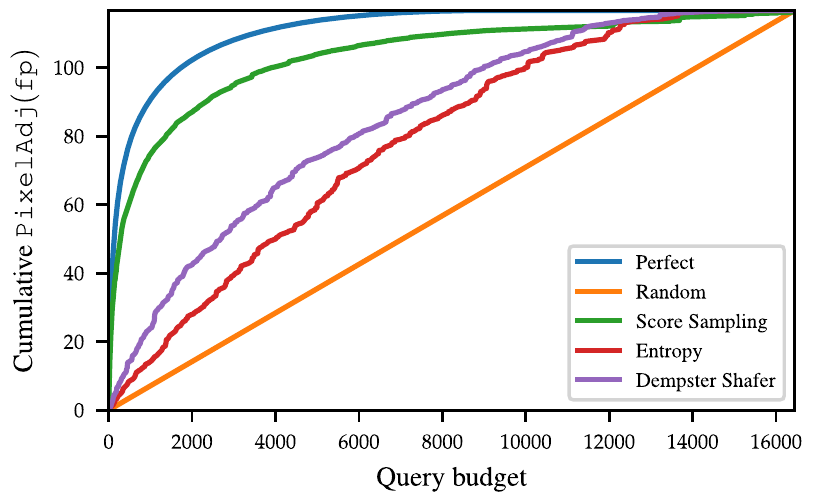}  
    \caption{Cumulative ground-truth hardness of boxes in images obtained with a fixed query budget for the overlap-adjusted hardness definition for \emph{mmdet-maskrcnn}.}
   \label{fig:regret_4}
\end{figure*}

In Figures~\ref{fig:regret_1}-\ref{fig:regret_4}, we show the cumulative ground-truth hardness for various hardness definitions for various detectors and see that score sampling finds these hard boxes much quicker than the baselines.

\section{Comparison of Hardness Definitions}
\label{sec:comparison_hardness_defs}
\begin{figure*}
    \centering
    \includegraphics[width=\linewidth]{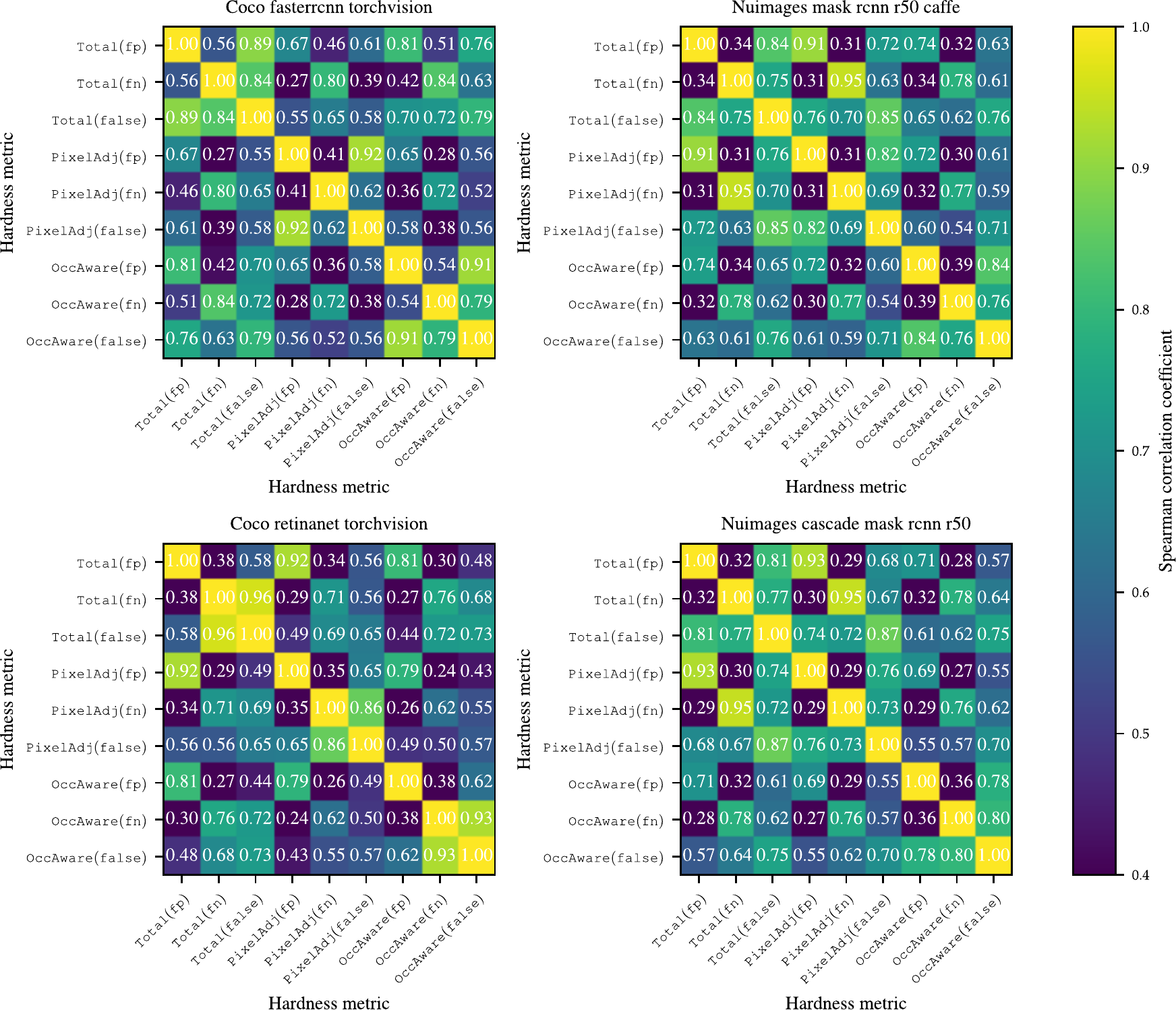}
    \caption{Heatmap of Spearman's rank correlation coefficient for RetinaNet and faster RCNN on the nuImages and COCO datasets.}
    \label{fig:hardness_correlation}
\end{figure*}

In Figure~\ref{fig:hardness_correlation} we show the correlation between the proposed hardness measures from the query based hardness section on the nuImages and COCO datasets for faster RCNN and RetinaNet detectors.
We observe that in general hardness measures may differ greatly in their ranking of the images in a dataset (i.e. false positive and false negative based metrics), although there are some similarities, e.g. hardness definitions which are re-weighted versions of another hardness definition.

\section{Mapping of Classes Between Datasets}
\label{sec:class mapping}
We remap the MMDet nuImages class labels to a coarser classification which only distinguishes between pedestrians and vehicles.
This enables a more salient evaluation of the ability of a detector to identify and distinguish between objects in a way that is important for autonomous driving.
It also means that we can compare with other AV datasets such as KITTI.
Objects which cannot be remapped to vehicles or pedestrians are discarded.
The class mapping is shown in Table~\ref{tab:class_remap}.

\begin{table}
    \centering
    \begin{tabular}{@{}l@{\hspace{4pt}}l@{\hspace{4pt}}}
    \toprule
        Original class & Remapped class  \\
        \midrule
        car & Vehicle \\
        truck & Vehicle \\
        trailer & Vehicle \\
        bus & Vehicle \\
        construction vehicle & Vehicle \\
        bicycle & Vehicle \\
        motorcycle & Vehicle \\
        pedestrian & Pedestrian \\
        traffic cone & None \\
        barrier & None \\
        \bottomrule
    \end{tabular}
    
    \caption{Class remapping from nuImages MMDet schema to simplified schema.}
    \label{tab:class_remap}
\end{table}

\section{Sensitivity to Number of Samples}
\label{sec:sensitivity}
\begin{figure*}
    \centering
    \subfigure[NDCG on \emph{coco-rcnn}]{ \includegraphics[width=0.45\linewidth]{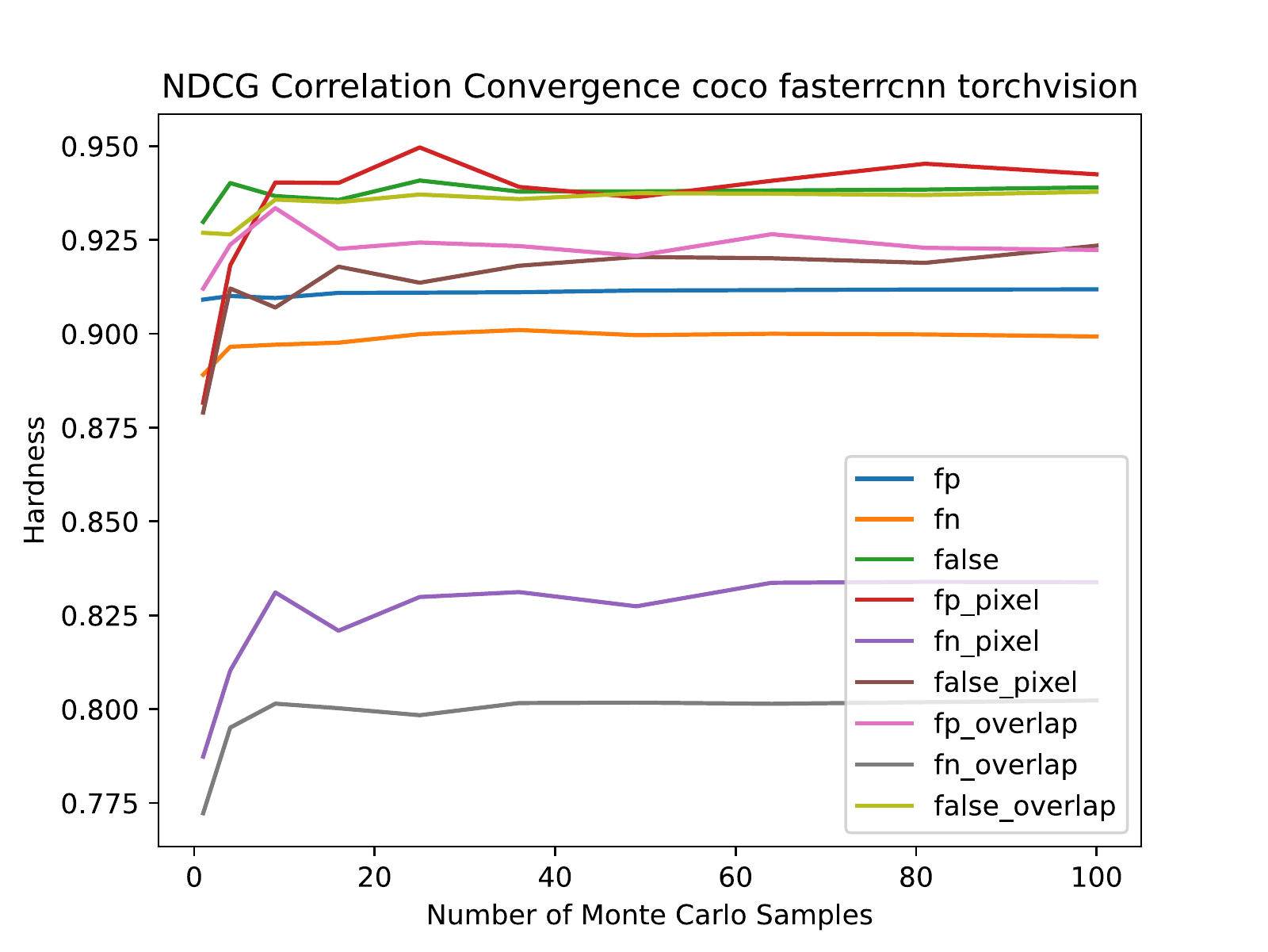}}
    \subfigure[NDCG on \emph{coco-retina}]{\includegraphics[width=0.45\linewidth]{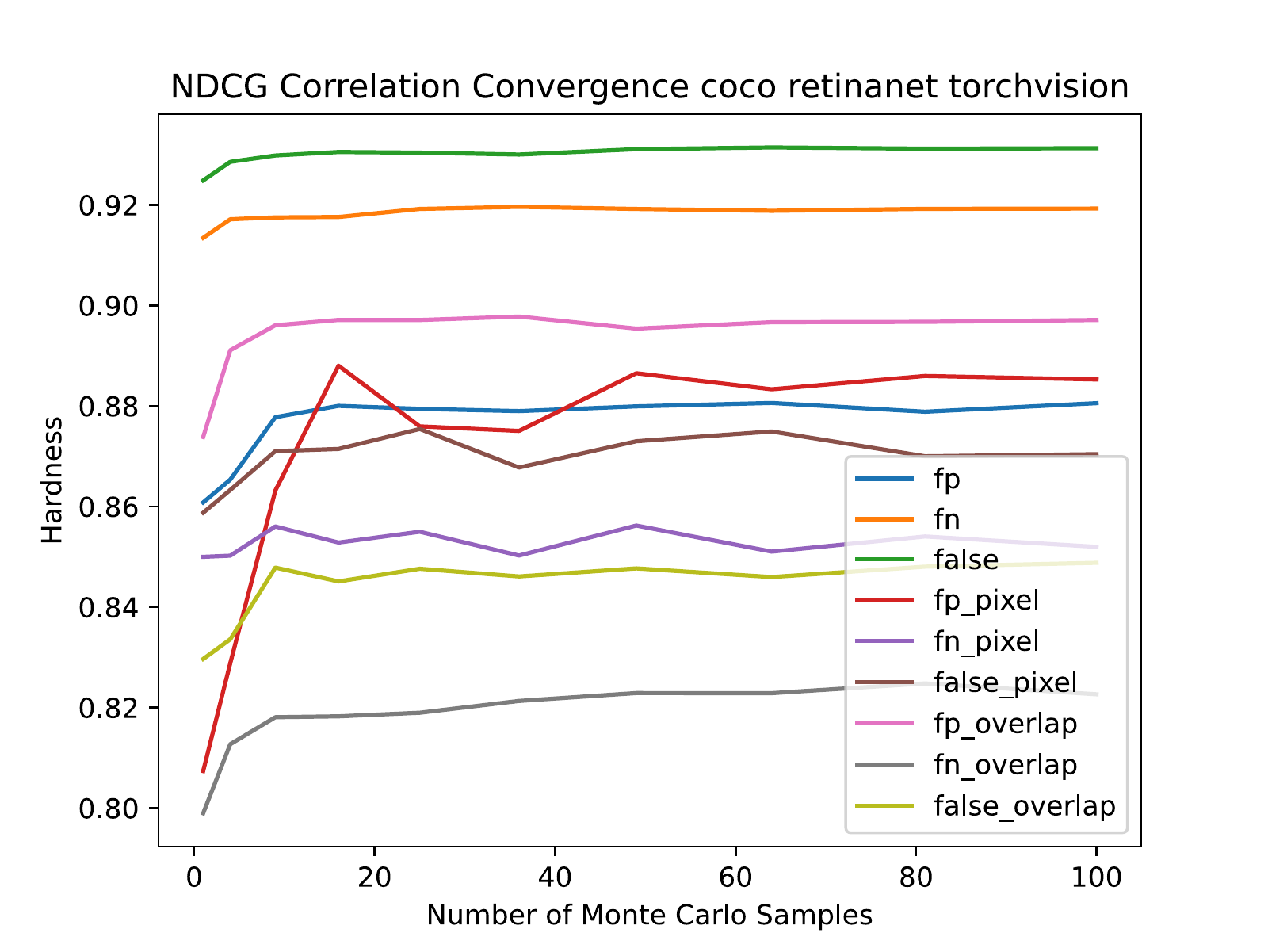}}\\
    \subfigure[ROC AUC on \emph{coco-rcnn}]{ \includegraphics[width=0.45\linewidth]{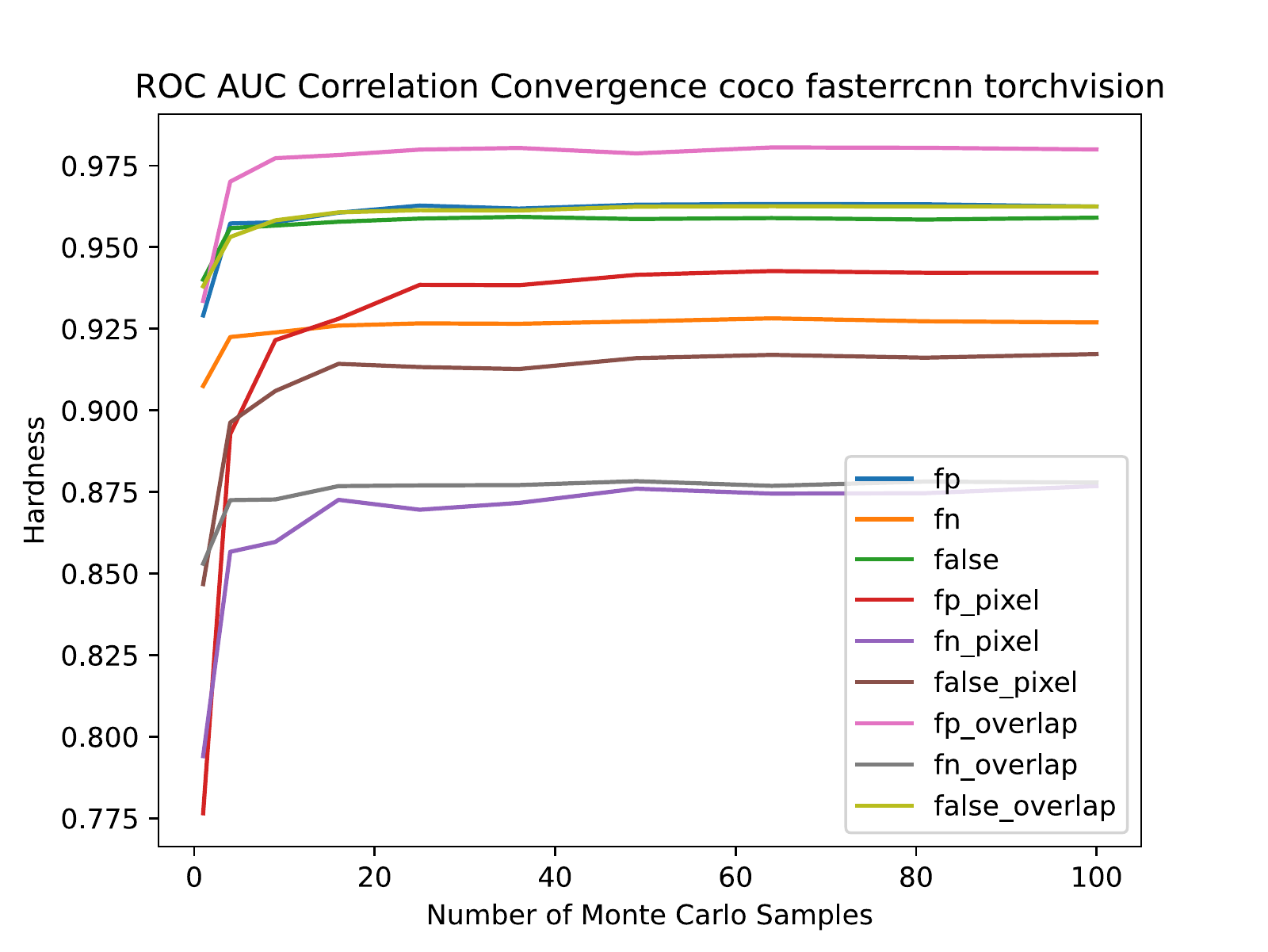}}
    \subfigure[ROC AUC on \emph{coco-retina}]{\includegraphics[width=0.45\linewidth]{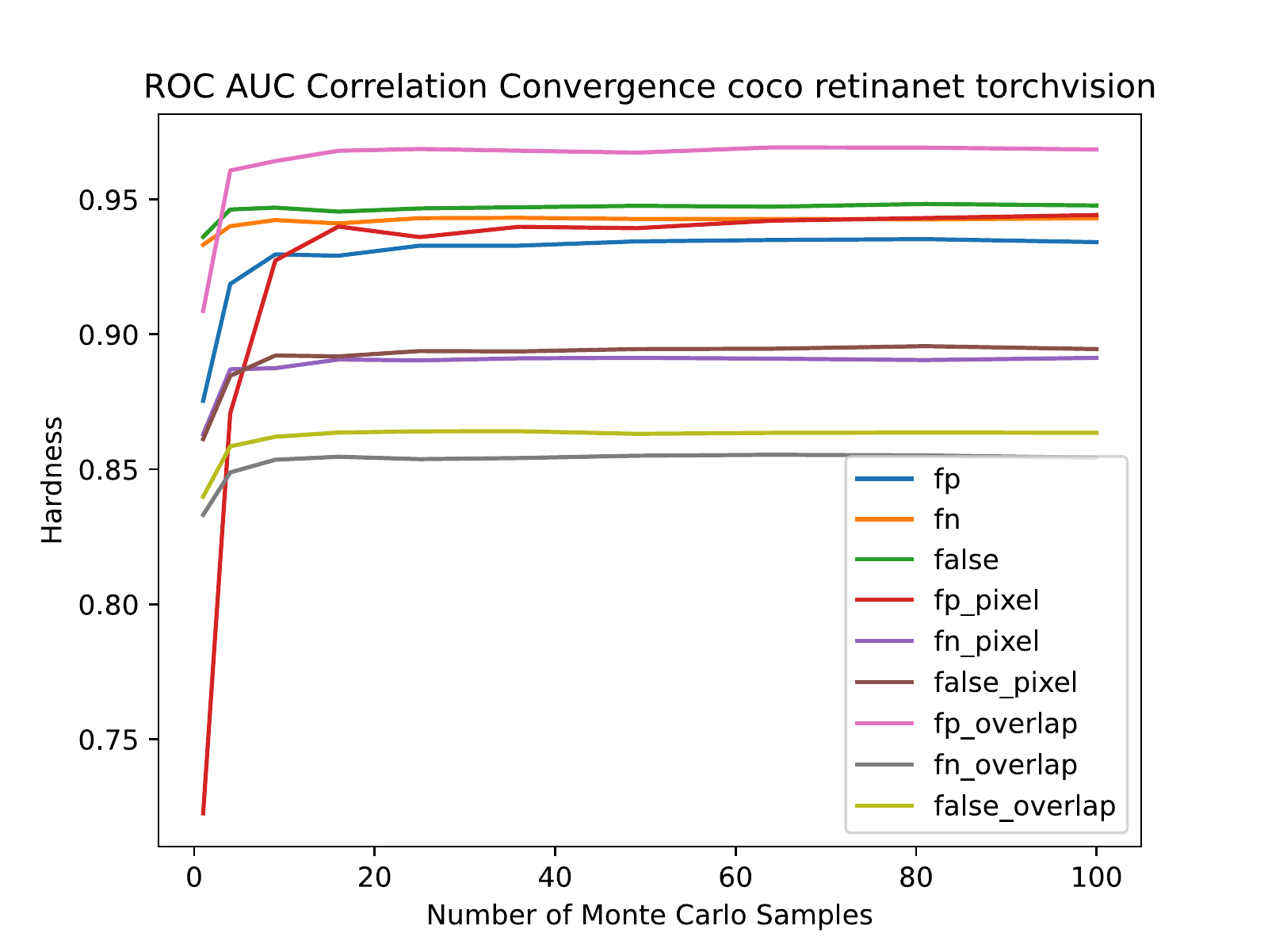}}
    \caption{Sensitivity of various evaluation metrics on \emph{coco-rcnn} and \emph{coco-retina} to number of Monte Carlo samples.}
    \label{fig:sensitivity_1}
\end{figure*}

Figure~\ref{fig:sensitivity_1} shows how the evaluation metrics for the ranking change with respect to the number of Monte Carlo score samples.
For a particular image the error in the expected hardness due to Monte Carlo approximation will decrease as $\frac{1}{\sqrt{N}}$, where $N$ is the number of Monte Carlo samples, since the formula for standard error in the mean is well known.
We see that for most metrics the evaluation metric stabilises after about 10 Monte Carlo samples, and hence we consider this an appropriate number of samples to use for our evaluation in the experiments, to balance speed and efficacy.

\section{Hardest Images} \label{sec:hardest}

\begin{figure*}[ht!]
\begin{tabularx}{\textwidth}{ c c c c }
\toprule
   & \multicolumn{3}{c}{\textbf{Query 1}: Total False Positives ($\texttt{Total}(\texttt{fp})$)} \\
 \cmidrule{2-4} 
 \raisebox{0.3cm}{\textbf{SS (our)}} & \includegraphics[height=2.6em]{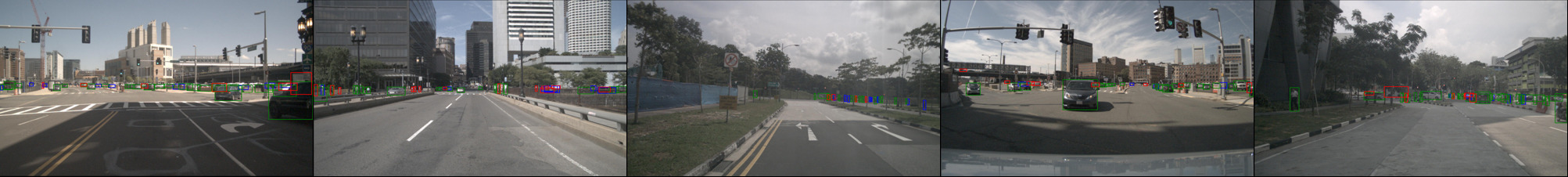} & \raisebox{0.3cm}{$\cdots$} & \includegraphics[height=2.6em]{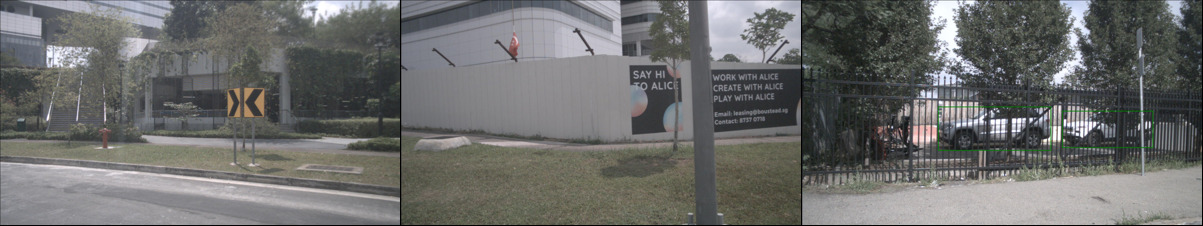} \\
 \raisebox{0.3cm}{\textbf{GT Ranking}} & \includegraphics[height=2.6em]{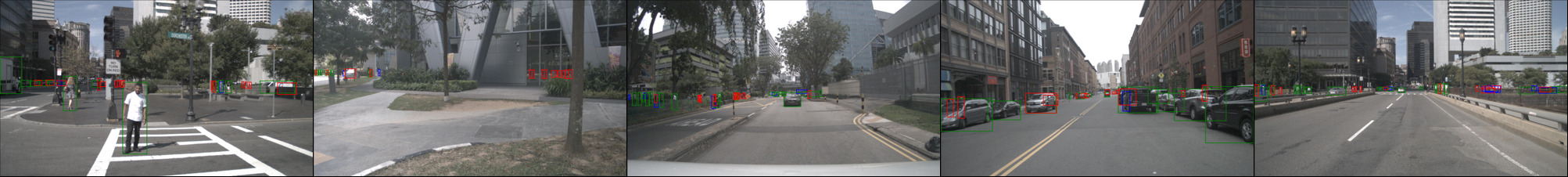} & \raisebox{0.3cm}{$\cdots$} & \includegraphics[height=2.6em]{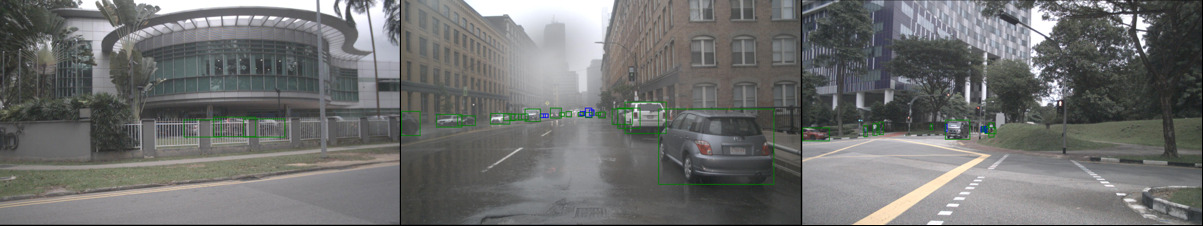} \\
 \cmidrule{2-4} 
   & \multicolumn{3}{c}{\textbf{Query 2}: Occlusion-aware False Positives ($\texttt{OccAware}(\texttt{fp})$)} \\
 \cmidrule{2-4} 
 \raisebox{0.3cm}{\textbf{SS (our)}} & \includegraphics[height=2.6em]{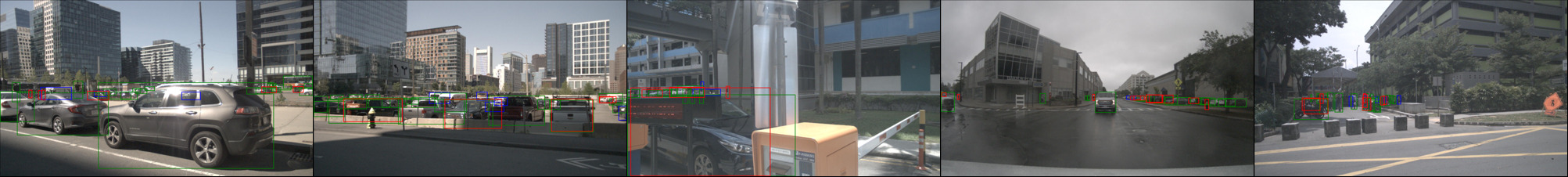} & \raisebox{0.3cm}{$\cdots$} & \includegraphics[height=2.6em]{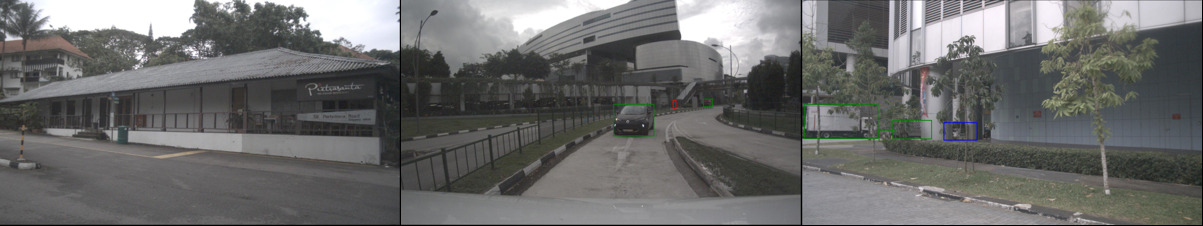} \\
 \raisebox{0.3cm}{\textbf{GT Ranking}} & \includegraphics[height=2.6em]{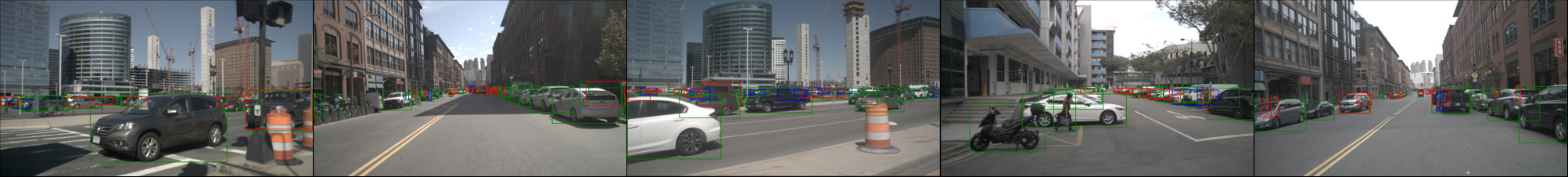} & \raisebox{0.3cm}{$\cdots$} & \includegraphics[height=2.6em]{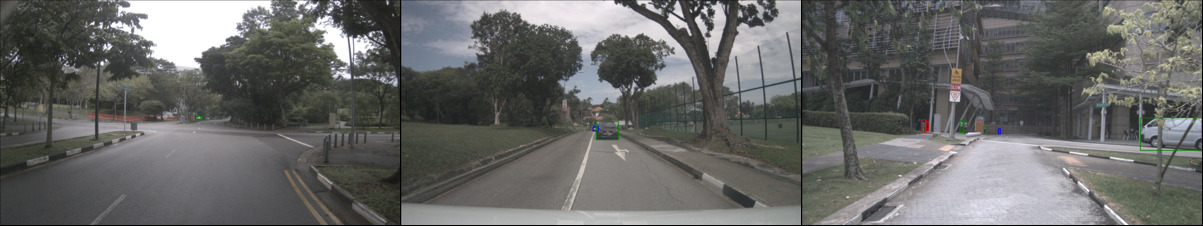} \\
\cmidrule{2-4} 
   & \multicolumn{3}{c}{\textbf{Query 3}: Pixel-adjusted False Positives ($\texttt{PixelAdj}(\texttt{fp})$)} \\
 \cmidrule{2-4} 
 \raisebox{0.3cm}{\textbf{SS (our)}} & \includegraphics[height=2.6em]{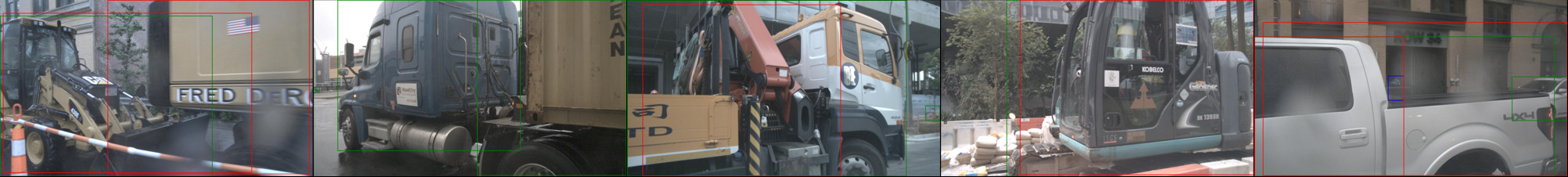} & \raisebox{0.3cm}{$\cdots$} & \includegraphics[height=2.6em]{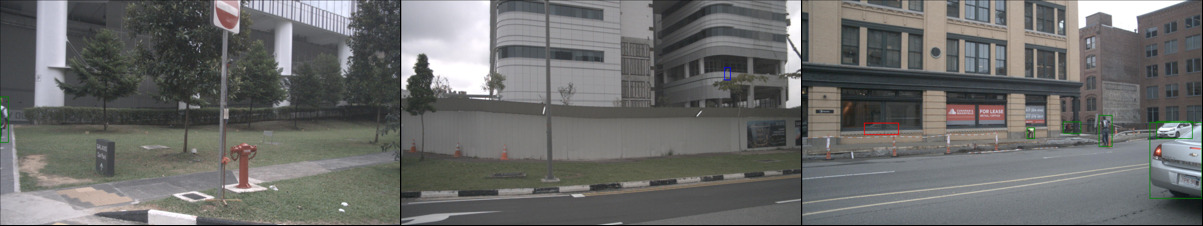} \\
 \raisebox{0.3cm}{\textbf{GT Ranking}} & \includegraphics[height=2.6em]{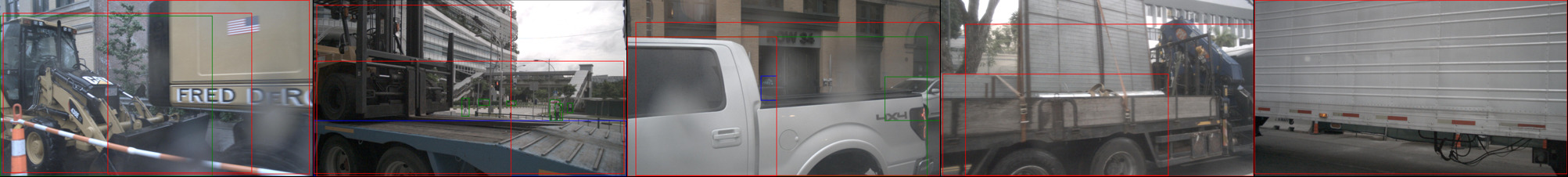} & \raisebox{0.3cm}{$\cdots$} & \includegraphics[height=2.6em]{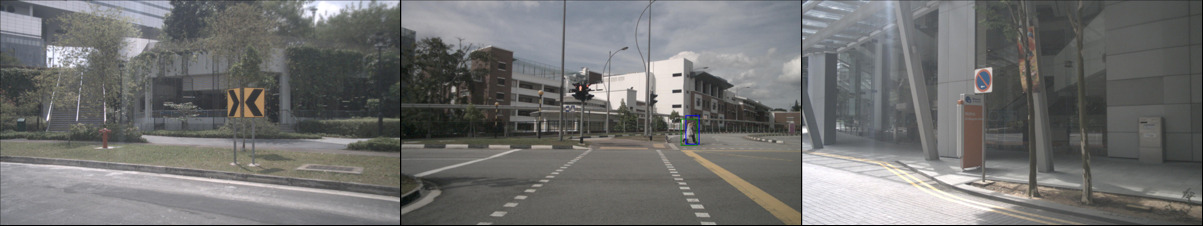} \\
 \cmidrule{2-4} 
 & \multicolumn{3}{c}{\textbf{\textcolor{red}{Query-agnostic Baselines}}} \\
 \cmidrule{2-4} 
 \raisebox{0.3cm}{\textbf{Entropy}} & \includegraphics[height=2.6em]{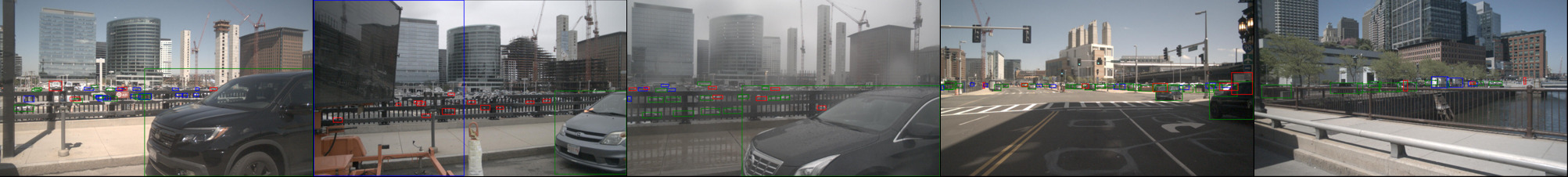} & \raisebox{0.3cm}{$\cdots$} & \includegraphics[height=2.6em]{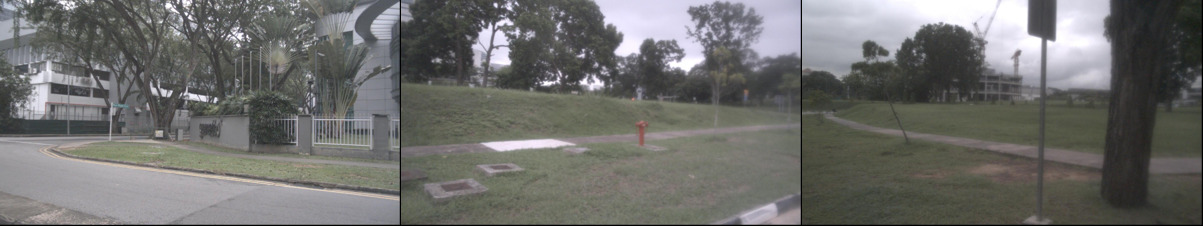} \\
 \raisebox{0.3cm}{\textbf{Dempster-Shafer}} & \includegraphics[height=2.6em]{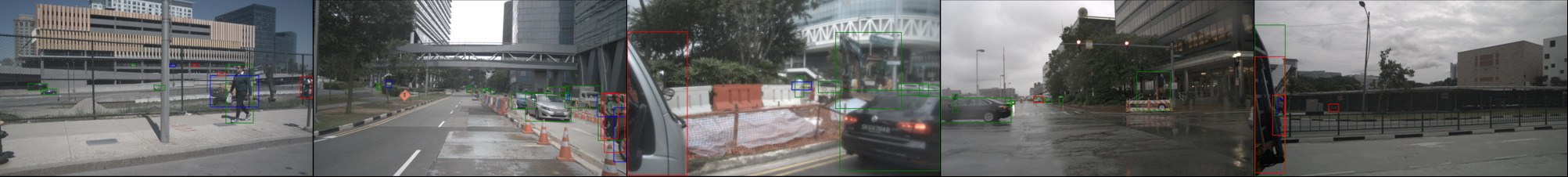} & \raisebox{0.3cm}{$\cdots$} & \includegraphics[height=2.6em]{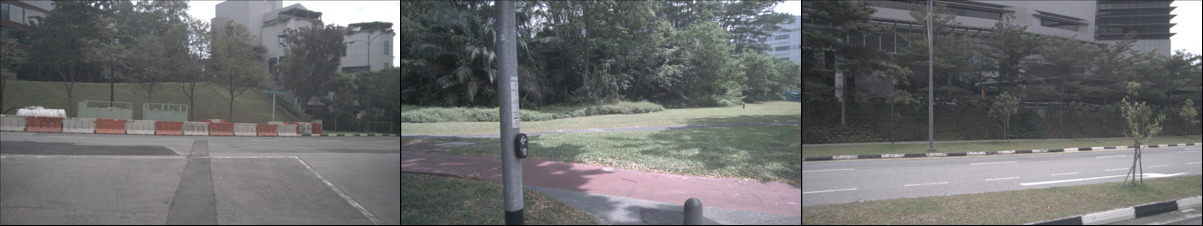} \\
\bottomrule
\end{tabularx}
\caption{Hardest images (left) and easiest images (right) for the \emph{mmdet-maskrcnn} on the nuImages \emph{val} dataset for different hardness queries.
Entropy and evidential measures of uncertainty are unable to adapt to different hardness queries and are hence query agnostic.
\textcolor{green}{Green}, \textcolor{red}{red}, and \textcolor{blue}{blue} boxes denote true positives, false positives, false negatives, respectively.
\label{fig:hardest_nuimages}}
\end{figure*}

\begin{figure*}
     \centering
\subfigure[Score Sampling]{\includegraphics[width=0.4\textwidth]{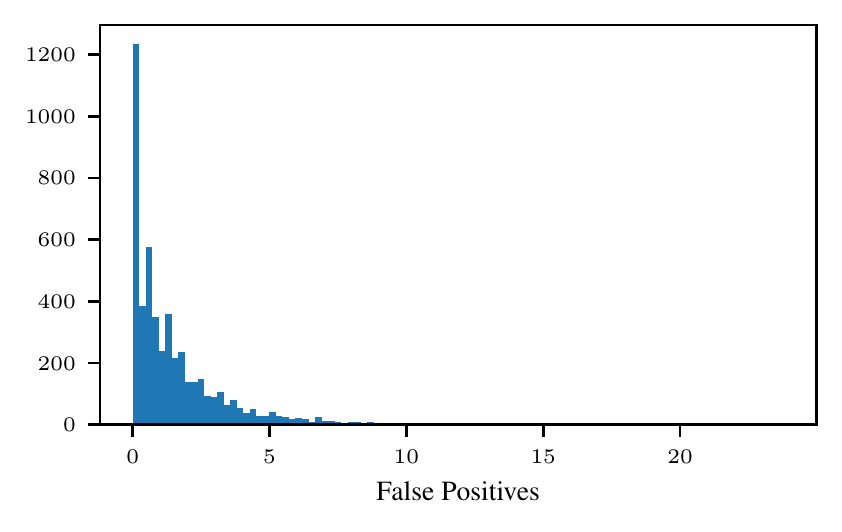}}
\subfigure[Ground Truth]{\includegraphics[width=0.4\textwidth]{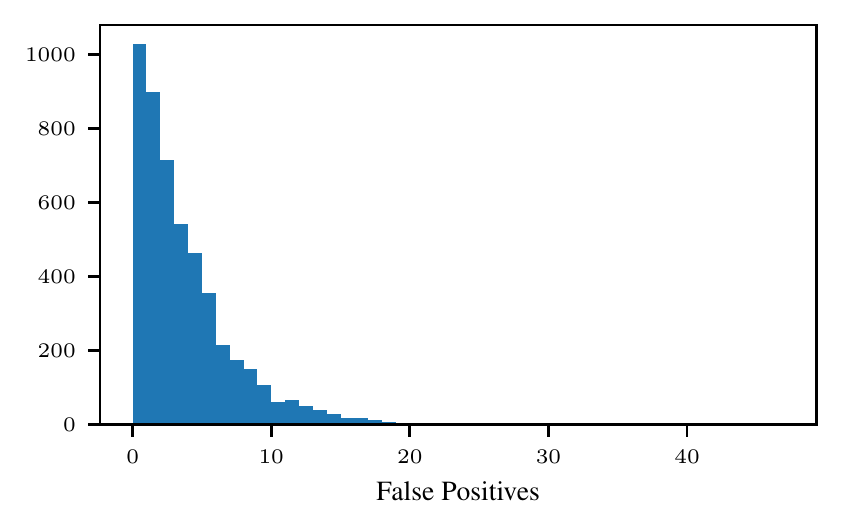}}\\
\subfigure[Entropy]{\includegraphics[width=0.4\textwidth]{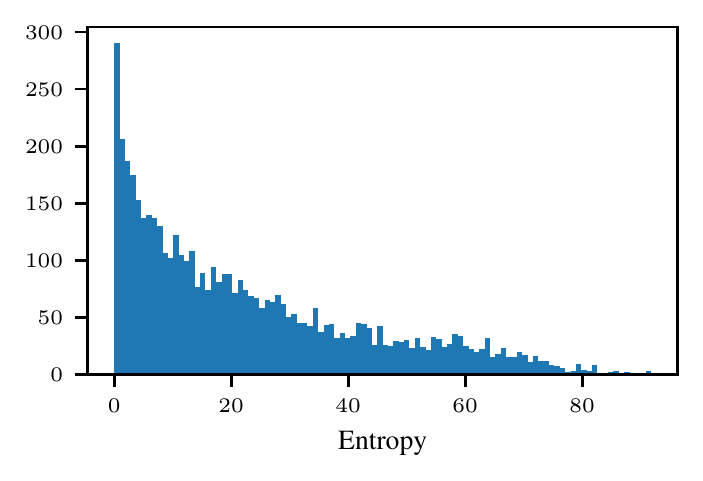}}
\subfigure[Evidential]{\includegraphics[width=0.4\textwidth]{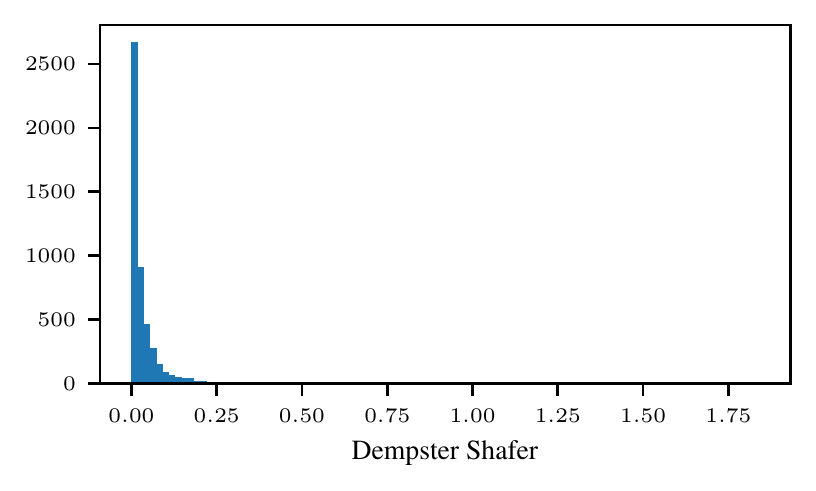}}
    \caption{Histograms of the estimated and actual number of false positives for \emph{coco-rcnn} on the COCO \emph{val} set. Note the large number of images with zero false positives.}
    \label{fig:histograms}
\end{figure*}

In Figure~\ref{fig:hardest_nuimages} we repeat the qualitative analysis for \emph{mmdet-maskrcnn} and show examples of the hardest images identified by the methods studied in the experiments section.
The images found by score sampling often are more qualitatively similar to the true hardest images than those obtained from the baselines.
Figure~\ref{fig:histograms} shows histograms for the estimated and actual number of false positives for \emph{coco-rcnn}.
Due to the large number of images with zero hardness there is no ranking for the easiest images, which should all be considered equally easy by this definition.

\section{Variance of Detected Scores} \label{sec:variance}

\begin{figure*}
     \centering
\subfigure[\emph{coco-rcnn}]{\includegraphics[width=0.4\textwidth]{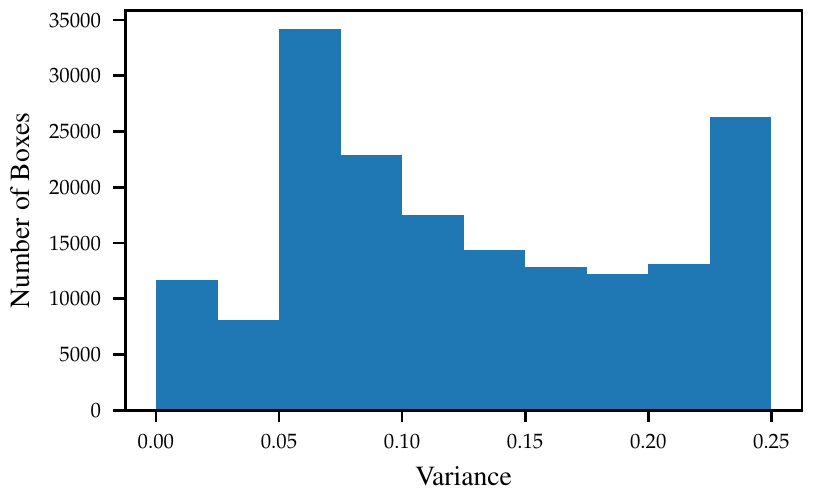}}
\subfigure[\emph{coco-retina}]{\includegraphics[width=0.4\textwidth]{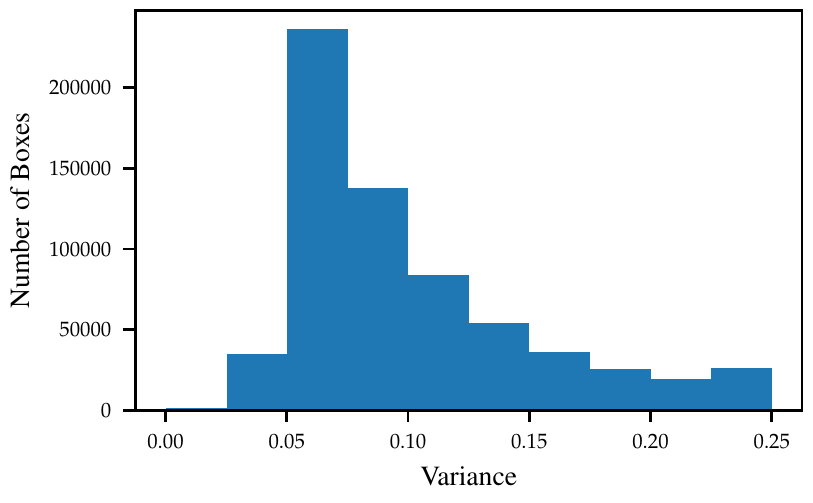}}\\
\subfigure[\emph{mmdet-maskrcnn}]{\includegraphics[width=0.4\textwidth]{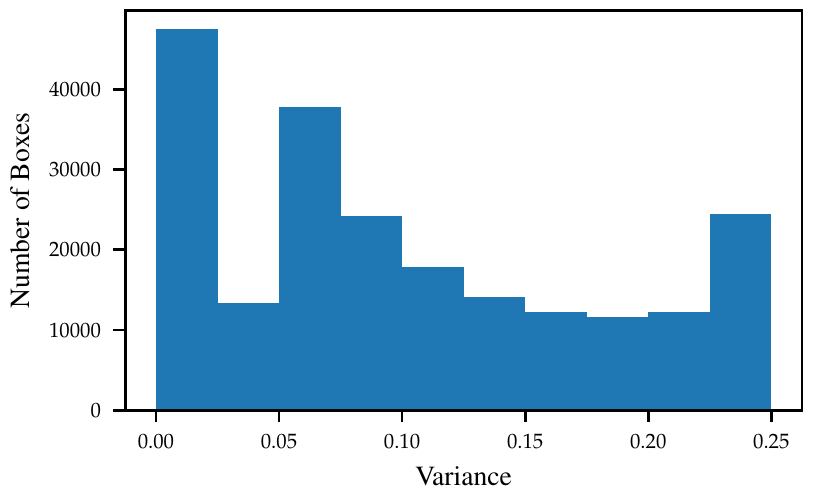}}
\subfigure[\emph{mmdet-cascade}]{\includegraphics[width=0.4\textwidth]{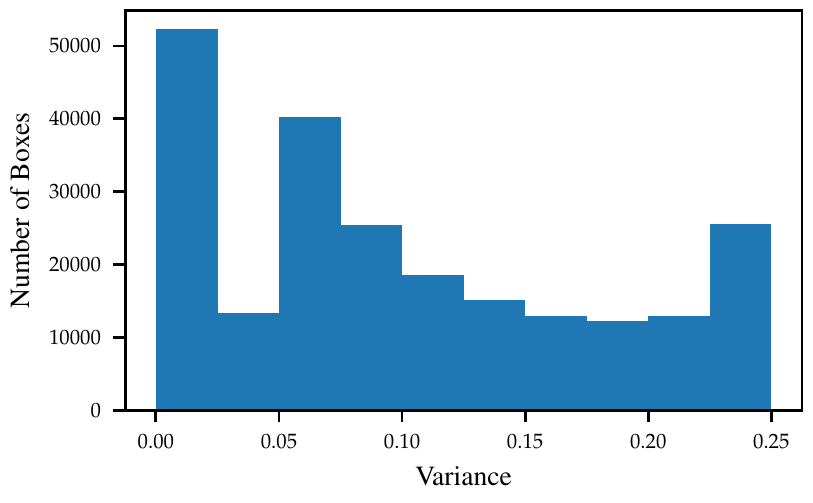}}
    \caption{Histograms of the score variances on the detectors and datasets used in this work.}
    \label{fig:variance_hist}
\end{figure*}

Figure~\ref{fig:variance_hist} shows the histograms of the variances of the Bernoulli random variables used for sampling pseudo ground-truth boxes from the detection scores (computed as $s_i(1-s_i)$) on the detectors and datasets used in this work. 
The histogram spreads from 0 to 0.25 and we observe that there is a significant mass over non-zero variances (e.g. 30\% of the detections have a variance above 0.17 for FasterRCNN on COCO test), indicating that different Monte Carlo samples are able to represent different configurations.

%% file: anonymous-submission-latex-2023.bbl
\begin{thebibliography}{30}
\providecommand{\natexlab}[1]{#1}

\bibitem[{Aghdam et~al.(2019)Aghdam, Gonzalez-Garcia, Weijer, and
  L{\'o}pez}]{Aghdam2019-dx}
Aghdam, H.~H.; Gonzalez-Garcia, A.; Weijer, J. v.~d.; and L{\'o}pez, A.~M.
  2019.
\newblock Active learning for deep detection neural networks.
\newblock In \emph{Proceedings of the {IEEE/CVF} International Conference on
  Computer Vision}, 3672--3680. openaccess.thecvf.com.

\bibitem[{Bucher, Herbin, and Jurie(2016)}]{Bucher2016-mp}
Bucher, M.; Herbin, S.; and Jurie, F. 2016.
\newblock Hard Negative Mining for Metric Learning Based {Zero-Shot}
  Classification.
\newblock In \emph{Computer Vision -- {ECCV} 2016 Workshops}, 524--531.
  Springer International Publishing.

\bibitem[{Caesar et~al.(2019)Caesar, Bankiti, Lang, Vora, Liong, Xu, Krishnan,
  Pan, Baldan, and Beijbom}]{nuscenes2019}
Caesar, H.; Bankiti, V.; Lang, A.~H.; Vora, S.; Liong, V.~E.; Xu, Q.; Krishnan,
  A.; Pan, Y.; Baldan, G.; and Beijbom, O. 2019.
\newblock nuScenes: A multimodal dataset for autonomous driving.
\newblock \emph{arXiv preprint arXiv:1903.11027}.

\bibitem[{Cai and Vasconcelos(2019)}]{cai2019cascade}
Cai, Z.; and Vasconcelos, N. 2019.
\newblock Cascade R-CNN: high quality object detection and instance
  segmentation.
\newblock \emph{IEEE transactions on pattern analysis and machine
  intelligence}, 43(5): 1483--1498.

\bibitem[{Chen et~al.(2020)Chen, Luo, Wu, Chen, and Peng}]{ovsampler}
Chen, J.; Luo, B.; Wu, Q.; Chen, J.; and Peng, X. 2020.
\newblock Overlap Sampler for Region-Based Object Detection.
\newblock In \emph{{WACV}}, 756--764.

\bibitem[{Chen et~al.(2019)Chen, Wang, Pang, Cao, Xiong, Li, Sun, Feng, Liu,
  Xu, Zhang, Cheng, Zhu, Cheng, Zhao, Li, Lu, Zhu, Wu, Dai, Wang, Shi, Ouyang,
  Loy, and Lin}]{mmdet3d}
Chen, K.; Wang, J.; Pang, J.; Cao, Y.; Xiong, Y.; Li, X.; Sun, S.; Feng, W.;
  Liu, Z.; Xu, J.; Zhang, Z.; Cheng, D.; Zhu, C.; Cheng, T.; Zhao, Q.; Li, B.;
  Lu, X.; Zhu, R.; Wu, Y.; Dai, J.; Wang, J.; Shi, J.; Ouyang, W.; Loy, C.~C.;
  and Lin, D. 2019.
\newblock {MMDetection}: Open MMLab Detection Toolbox and Benchmark.
\newblock \emph{arXiv preprint arXiv:1906.07155}.

\bibitem[{Choi et~al.(2021)Choi, Elezi, Lee, Farabet, and
  Alvarez}]{Choi2021-zg}
Choi, J.; Elezi, I.; Lee, H.-J.; Farabet, C.; and Alvarez, J.~M. 2021.
\newblock Active learning for deep object detection via probabilistic modeling.
\newblock In \emph{Proceedings of the IEEE/CVF International Conference on
  Computer Vision}, 10264--10273.

\bibitem[{Dempster(2008)}]{Dempster2008}
Dempster, A.~P. 2008.
\newblock A generalization of Bayesian inference.
\newblock In \emph{Classic works of the Dempster-Shafer theory of belief
  functions}.

\bibitem[{Du et~al.(2022)Du, Wang, Cai, and Li}]{VOS}
Du, X.; Wang, Z.; Cai, M.; and Li, S. 2022.
\newblock Towards Unknown-aware Learning with Virtual Outlier Synthesis.
\newblock In \emph{International Conference on Learning Representations}.

\bibitem[{Feng et~al.(2019)Feng, Wei, Rosenbaum, Maki, and
  Dietmayer}]{Feng2019-yt}
Feng, D.; Wei, X.; Rosenbaum, L.; Maki, A.; and Dietmayer, K. 2019.
\newblock Deep active learning for efficient training of a lidar 3d object
  detector.
\newblock In \emph{2019 IEEE Intelligent Vehicles Symposium (IV)}, 667--674.
  IEEE.

\bibitem[{Forsyth and Ponce(2012)}]{forsyth2012computer}
Forsyth, D.~A.; and Ponce, J. 2012.
\newblock \emph{Computer vision: a modern approach}.
\newblock Pearson.

\bibitem[{Gal, Islam, and Ghahramani(2017)}]{Gal2017-vh}
Gal, Y.; Islam, R.; and Ghahramani, Z. 2017.
\newblock Deep bayesian active learning with image data.
\newblock In \emph{International Conference on Machine Learning}, 1183--1192.
  proceedings.mlr.press.

\bibitem[{Harakeh and Waslander(2021)}]{RegressionUncOD}
Harakeh, A.; and Waslander, S.~L. 2021.
\newblock Estimating and Evaluating Regression Predictive Uncertainty in Deep
  Object Detectors.
\newblock In \emph{International Conference on Learning Representations
  (ICLR)}.

\bibitem[{Haussmann et~al.(2020)Haussmann, Fenzi, Chitta, Ivanecky, Xu, Roy,
  Mittel, Koumchatzky, Farabet, and Alvarez}]{Haussmann2020-vg}
Haussmann, E.; Fenzi, M.; Chitta, K.; Ivanecky, J.; Xu, H.; Roy, D.; Mittel,
  A.; Koumchatzky, N.; Farabet, C.; and Alvarez, J.~M. 2020.
\newblock Scalable active learning for object detection.
\newblock In \emph{2020 IEEE intelligent vehicles symposium (iv)}, 1430--1435.
  IEEE.

\bibitem[{He et~al.(2017)He, Gkioxari, Doll{\'a}r, and Girshick}]{he2017mask}
He, K.; Gkioxari, G.; Doll{\'a}r, P.; and Girshick, R. 2017.
\newblock Mask r-cnn.
\newblock In \emph{Proceedings of the IEEE international conference on computer
  vision}, 2961--2969.

\bibitem[{J{\"a}rvelin and Kek{\"a}l{\"a}inen(2002)}]{jarvelin2002cumulated}
J{\"a}rvelin, K.; and Kek{\"a}l{\"a}inen, J. 2002.
\newblock Cumulated gain-based evaluation of IR techniques.
\newblock \emph{ACM Transactions on Information Systems (TOIS)}, 20(4):
  422--446.

\bibitem[{Lin et~al.(2017)Lin, Goyal, Girshick, He, and Doll{\'a}r}]{retinanet}
Lin, T.-Y.; Goyal, P.; Girshick, R.; He, K.; and Doll{\'a}r, P. 2017.
\newblock Focal loss for dense object detection.
\newblock In \emph{Proceedings of the {IEEE} international conference on
  computer vision}, 2980--2988. openaccess.thecvf.com.

\bibitem[{Lin et~al.(2014)Lin, Maire, Belongie, Hays, Perona, Ramanan,
  Doll{\'a}r, and Zitnick}]{lin2014microsoft}
Lin, T.-Y.; Maire, M.; Belongie, S.; Hays, J.; Perona, P.; Ramanan, D.;
  Doll{\'a}r, P.; and Zitnick, C.~L. 2014.
\newblock Microsoft coco: Common objects in context.
\newblock In \emph{European conference on computer vision}, 740--755. Springer.

\bibitem[{Liu et~al.(2020)Liu, Lin, Padhy, Tran, Bedrax-Weiss, and
  Lakshminarayanan}]{Liu2020-sngp}
Liu, J.~Z.; Lin, Z.; Padhy, S.; Tran, D.; Bedrax-Weiss, T.; and
  Lakshminarayanan, B. 2020.
\newblock Simple and Principled Uncertainty Estimation with Deterministic Deep
  Learning via Distance Awareness.
\newblock In \emph{NeurIPS}.

\bibitem[{Mukhoti et~al.(2020)Mukhoti, Kulharia, Sanyal, Golodetz, Torr, and
  Dokania}]{mukhoti2020calibrating}
Mukhoti, J.; Kulharia, V.; Sanyal, A.; Golodetz, S.; Torr, P.~H.; and Dokania,
  P.~K. 2020.
\newblock Calibrating Deep Neural Networks using Focal Loss.
\newblock In \emph{NeurIPS}.

\bibitem[{Pang et~al.(2019)Pang, Chen, Shi, Feng, Ouyang, and Lin}]{LibraRCNN}
Pang, J.; Chen, K.; Shi, J.; Feng, H.; Ouyang, W.; and Lin, D. 2019.
\newblock Libra {R-CNN:} {T}owards Balanced Learning for Object Detection.
\newblock In \emph{The IEEE Conference on Computer Vision and Pattern
  Recognition (CVPR)}.

\bibitem[{Paszke et~al.(2019)Paszke, Gross, Massa, Lerer, Bradbury, Chanan,
  Killeen, Lin, Gimelshein, Antiga et~al.}]{paszke2019pytorch}
Paszke, A.; Gross, S.; Massa, F.; Lerer, A.; Bradbury, J.; Chanan, G.; Killeen,
  T.; Lin, Z.; Gimelshein, N.; Antiga, L.; et~al. 2019.
\newblock Pytorch: An imperative style, high-performance deep learning library.
\newblock \emph{Advances in neural information processing systems}, 32.

\bibitem[{Pinto et~al.(2022)Pinto, Yang, Lim, Torr, and
  Dokania}]{PintoRegMixup2022}
Pinto, F.; Yang, H.; Lim, S.-N.; Torr, P.~H.; and Dokania, P.~K. 2022.
\newblock RegMixup: Mixup as a Regularizer Can Surprisingly Improve Accuracy
  and Out Distribution Robustness.
\newblock In \emph{arXiv: 2206.14502}.

\bibitem[{Ren et~al.(2015)Ren, He, Girshick, and Sun}]{ren2015faster}
Ren, S.; He, K.; Girshick, R.; and Sun, J. 2015.
\newblock Faster r-cnn: Towards real-time object detection with region proposal
  networks.
\newblock \emph{Advances in neural information processing systems}, 28.

\bibitem[{Roy, Unmesh, and Namboodiri(2018)}]{Roy2018-lv}
Roy, S.; Unmesh, A.; and Namboodiri, V.~P. 2018.
\newblock Deep active learning for object detection.
\newblock In \emph{{BMVC}}, volume 362, 91. bmva.org.

\bibitem[{Sensoy, Kaplan, and Kandemir(2018)}]{sensoy2018evidential}
Sensoy, M.; Kaplan, L.; and Kandemir, M. 2018.
\newblock Evidential deep learning to quantify classification uncertainty.
\newblock \emph{Advances in Neural Information Processing Systems}, 31.

\bibitem[{Shannon(1948)}]{ShannonEntropy}
Shannon, C.~E. 1948.
\newblock A mathematical theory of communication.
\newblock In \emph{Bell System Technical Journal}.

\bibitem[{Shrivastava, Gupta, and Girshick(2016)}]{OHEM}
Shrivastava, A.; Gupta, A.; and Girshick, R. 2016.
\newblock Training Region-based Object Detectors with Online Hard Example
  Mining.
\newblock In \emph{The IEEE Conference on Computer Vision and Pattern
  Recognition (CVPR)}.

\bibitem[{Suh et~al.(2019)Suh, Han, Kim, and Lee}]{Suh2019-tf}
Suh, Y.; Han, B.; Kim, W.; and Lee, K.~M. 2019.
\newblock Stochastic class-based hard example mining for deep metric learning.
\newblock In \emph{Proceedings of the {IEEE/CVF} Conference on Computer Vision
  and Pattern Recognition}, 7251--7259. openaccess.thecvf.com.

\bibitem[{Yu et~al.(2022)Yu, Zhu, Yang, and Chen}]{Yu2021-fe}
Yu, W.; Zhu, S.; Yang, T.; and Chen, C. 2022.
\newblock Consistency-based active learning for object detection.
\newblock In \emph{Proceedings of the IEEE/CVF Conference on Computer Vision
  and Pattern Recognition}, 3951--3960.

\end{thebibliography}
